\documentclass[10pt,twocolumn,letterpaper]{article}
\usepackage{iccv,times,graphicx,amsmath,amssymb,booktabs}
\usepackage{epsfig}
\usepackage{graphicx}
\usepackage{amsmath}
\usepackage{amssymb}

\usepackage{algorithm}
\usepackage{algorithmic}
\usepackage{subcaption}
\usepackage{booktabs}
\usepackage{appendix}
\iccvfinalcopy
\usepackage[pagebackref=true,breaklinks=true,letterpaper=true,colorlinks,bookmarks=false]{hyperref}

\def\iccvPaperID{4038} % *** Enter the ICCV Paper ID here

\ificcvfinal\fi

\begin{document}

%%%%%%%%% TITLE
\title{Architecture Disentanglement for Deep Neural Networks}

\author{
Jie Hu$^1$, Liujuan Cao$^1$, Qixiang Ye$^2$, Tong Tong$^1$, Shengchuan Zhang$^1$, \\Ke Li$^3$, Feiyue Huang$^3$, Rongrong Ji$^{1}$, and Ling Shao$^4$.\\
$^1$Xiamen University,
$^2$University of Chinese Academy of Sciences,\\
$^3$Tencent Youtu Lab,
$^4$Inception Institute of Artificial Intelligence.
}

\maketitle
\ificcvfinal\thispagestyle{empty}\fi

%%%%%%%%% ABSTRACT
\begin{abstract}
Understanding the inner workings of deep neural networks (DNNs) is essential to provide trustworthy artificial intelligence techniques for practical applications.
Existing studies typically involve linking semantic concepts to units or layers of DNNs, but fail to explain the inference process.
In this paper, we introduce neural architecture disentanglement (NAD) to fill the gap.
Specifically, NAD learns to disentangle a pre-trained DNN into sub-architectures according to independent tasks, forming information flows that describe the inference processes.
We investigate whether, where, and how the disentanglement occurs through experiments conducted with handcrafted and automatically-searched network architectures, on both object-based and scene-based datasets.
Based on the experimental results, we present three new findings that provide fresh insights into the inner logic of DNNs.
First, DNNs can be divided into sub-architectures for independent tasks.
Second, deeper layers do not always correspond to higher semantics.
Third, the connection type in a DNN affects how the information flows across layers, leading to different disentanglement behaviors.
With NAD, we further explain why DNNs sometimes give wrong predictions.
Experimental results show that misclassified images have a high probability of being assigned to task sub-architectures similar to the correct ones.
%
% Our code will be made available to facilitate future research.
%
\end{abstract}

%%%%%%%%% BODY TEXT
\section{Introduction}
% \label{Introduction}
%
A fundamental problem in using deep neural networks (DNNs) is our inability to understand their inner workings, which is crucial in many real-world applications, including healthcare, criminal justice, and administrative regulation~\cite{rudin2019stop}.
Thus, interpreting DNNs has attracted ever-increasing research attention in recent years.
Existing endeavors typically link semantics to units or layers of DNNs to determine the roles of these specific parts.
\begin{figure}[t]
 \centering
 \includegraphics[width=0.98\linewidth]{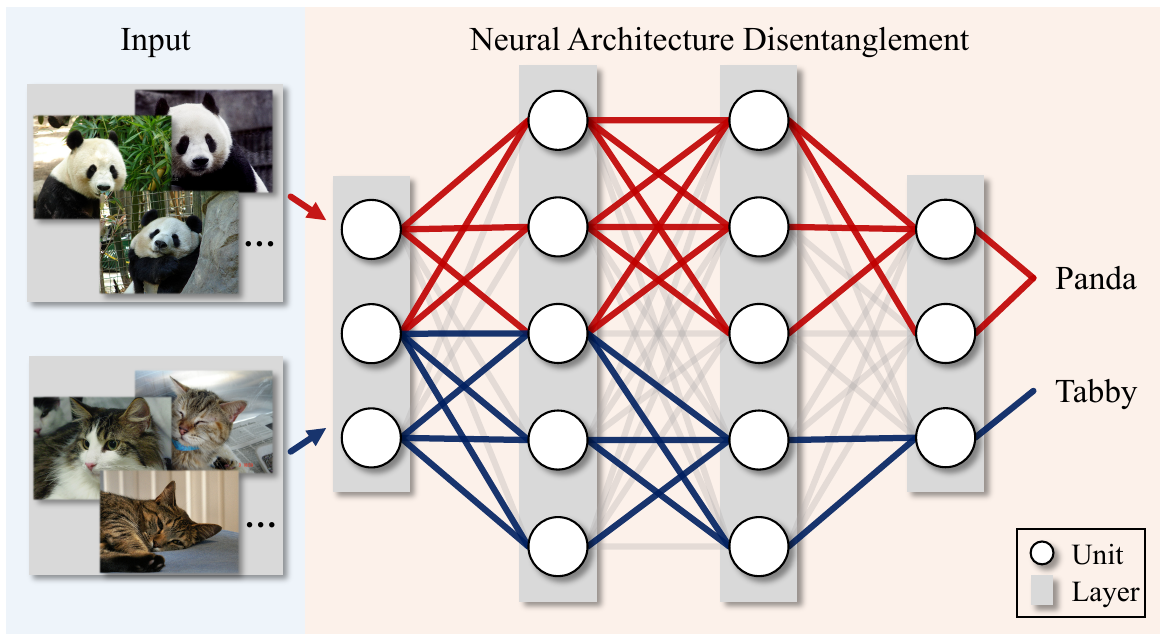}
 \caption{
 Illustration of the proposed neural architecture disentanglement (NAD).
 NAD aims to disentangle a pre-trained DNN into sub-architectures, each of which is responsible for only one task.
 For example, the network in this figure is disentangled into two sub-architectures, \ie, the red one and the blue one, for the task of classifying `Panda' and `Tabby'.
 Note that the sub-architectures can overlap with the same unit for different tasks.
 }%\vspace{-3mm}
 \label{fig1}
\end{figure}
However, the hierarchical inference process is not effectively captured in this way, for two reasons.
First, the trained networks entangle information together, and one unit can be responsible for multiple classes~\cite{s2018on, zhou2018revisiting}.
Second, only knowing which unit or layer represents what class is insufficient for understanding the reasoning process in DNNs.
The relationship between successive layers is not explored.
%
% For instance, the `conv5-3 unit 151' of VGG16~\cite{simonyan2014very} pre-trained on ImageNet~\cite{deng2009imagenet} was shown to relate to class `Airplane'~\cite{netdissect2017}, but it remains unclear how and why this unit infers such a class from the input image.
%
For example, if we want to explain the inference process for `Airplane', we know from previous works that the bottom layer activates `Blue Sky' and the top layer activates `Airplane', but this still fails to explain how DNNs infer from `Blue Sky' to `Airplane'.
%
% On the contrary, our method explores how the features of the previous layers are selected for the latter layers.
Thus, if the network architectures could be disentangled into the sub-architectures in terms of tasks, such as classifying `Airplane', the above concerns could be naturally addressed.
For example, we can explain that the edges and colors from bottom layers are clustered in middle layers to form the parts, such as `Wing' and `Window', of `Airplane', and then the most representative part will be selected to activate the class `Airplane'.
In this paper, we introduce a new method termed neural architecture disentanglement (NAD), which learns to disentangle a pre-trained DNN into sub-architectures for different tasks.
As illustrated in Fig.~\ref{fig1}, the sub-architectures form the information flows that describe the inference processes.
% the information-theory-based
Inspired by representation disentanglement, we design an objective function constraining the information between successive layers of DNNs.
The hidden units are selected to construct the sub-architectures from a DNN's bottom layers to its top layers.
Extensive experiments are conducted to investigate whether, where, and how the disentanglement occurs.
The architectures used in our experiments range from handcrafted to automatically-searched, \ie, VGG16~\cite{simonyan2014very}, ResNet50~\cite{he2016deep}, DenseNet121~\cite{huang2017densely} and DARTS-Net~\cite{liu2018darts}.
Consistent results are obtained on both object-based and scene-based datasets, \ie, ImageNet~\cite{deng2009imagenet} and Place365~\cite{zhou2017places}, yielding three new findings that provide fresh insights into the inner logic of DNNs.

% 1)
First, we provide evidence that DNNs can be divided according to independent tasks, \ie, DNNs can be disentangled.
We compare the classification results of the original architectures and the disentangled sub-architectures in Fig.~\ref{fig2}.
After disentanglement, the Top@1 classification accuracies are squashed into the interval of $(90, 100]$ from $(70, 90]$ on ImageNet and $(40, 70]$ on Place365, which indicates that the sub-architectures correlate to the assigned tasks.
% 2)
Second, we find that deeper layers do not necessarily correspond to higher semantics, \ie, the disentanglement can end before the last layer.
Previous studies~\cite{netdissect2017,zeiler2014visualizing} show that the bottom layers of DNNs extract low-level features (\eg, edges and colors), while the top layers extract high-level features (\eg, object parts) for classification.
%
% As such, it is natural to see the sub-architectures overlap in the bottom layers (\ie, when they share low-level information) and begin to disentangle in the middle layers (\ie, when they begin combining high-level information with low-level information).
%
Our new observation is that the disentanglement of high-level information can end before the last layer in architectures with skip-connections.
For example, as shown in Figs.~\ref{fig4-2} and ~\ref{fig4-4}, the top class hit rates of ResNet50~\cite{he2016deep} (which has plain skip-connections) and DARTS-Net~\cite{liu2018darts} (which has automatically-searched skip-connections) appear in the 16th and 15th layer, respectively, instead of the last layer.
% 3)
Third, the connection type in a DNN affects how the information flows across layers, leading to different disentanglement behaviors.
Intuitively, direct connections successively transmit information layer by layer, while skip-connections amortize the information over all layers.
This makes the overall inference processes behave differently in the architectures with direct connections and with skip-connections.
% which results in high information fusion to the inference processes
Importantly, dense skip-connections severely mix up the information for classification.
For example, as shown in Fig.~\ref{fig4-3}, the visualized feature maps of DenseNet121 activate less useful patterns in some layers. However, the high-level information can still be extracted from the last layer, suggesting that the valuable information for classification is amortized into each layer.
Furthermore, from the perspective of NAD, we provide an explanation for why DNNs sometimes give wrong predictions.
Experimental results show that misclassified images have a high probability of being assigned to tasks with similar sub-architectures to the correct ones.

In summary, the contributions of this study include:
\begin{itemize}
 \vspace{-5px}
 \item We propose a new method, termed neural architecture disentanglement (NAD), to understand the inference process of DNNs. Our approach is rooted in classical information theory with an objective function constraining the information between successive layers of DNNs.
 \vspace{-5px}
 \item We study the properties of NAD with network architectures ranging from handcrafted to automatically-searched. Consistent results on scene-based and object-based datasets yield three new findings for the inner workings of DNNs.
 \vspace{-5px}
\end{itemize}

%
% The rest of this paper is organized as follows:
% %
% Section~\ref{RW} reviews the related works.
% %
% The target and method of NAD are introduced in Section~\ref{Method}.
% %
% Experimental results are given in Section~\ref{Experiments}.
% %
% Finally, we conclude this work in Section~\ref{Conclusion}.
%

\section{Related Work}
\label{RW}
\noindent\textbf{Visualization-Based Interpretability.}
Visualization-based interpretability aims to reveal the interpretability of DNNs by visualization~\cite{erhan2009visualizing,simonyan2013deep,yosinski2015understanding,nguyen2016multifaceted,nguyen2016synthesizing,cadena2018diverse}.
Many studies have been conducted to explain DNNs by visualizing feature maps and then assigning semantic concepts to individual units or layers~\cite{gonzalez2018semantic,zeiler2014visualizing}.
For example, activation maximization~\cite{erhan2009visualizing,nguyen2019understanding} optimizes a randomly initialized input to maximize a specific unit and then assigns a concept to this unit by observing the optimized input.
Network dissection~\cite{netdissect2017, bau2019gandissect} aligns individual hidden units with a set of semantic concepts using predefined pixel-level labels.
The concept activation vector~\cite{kim2017interpretability} interprets the internal layers of DNNs in terms of concepts by learning a normal vector of a plane that differentiates the selected concept from others.
Concept whitening~\cite{chen2020concept} tries to align layers to the activated concepts using image classes.
However, the relationships between successive layers are not explored in the above methods.
%
% For example, if we want to explain the inference for the concept ``airplane'', we know from previous works that the bottom layer activates the concept ``blue sky'' and the top layer activates the concept ``airplane'', but this still fails to explain how DNNs infer from ``blue sky'' to ``airplane''.
%
In contrast, our method explores how the features of the previous layers are selected for the latter layers, thus showing the overall inference processes for classifying specific classes.
%
% For example, we can explain that the edges and colors from bottom layers are clustered in middle layers to form the parts such as ``wing'' and ``window'' of ``airplane'', and then the most related part will be selected to activate the concept ``airplane''.

\noindent\textbf{Disentanglement-Based Interpretability.} An interpretable DNN can also be trained by enforcing the disentanglement of representations~\cite{higgins2017beta, zhang2018interpretable,2021arXiv210311251R} or distilling DNNs into more interpretable models~\cite{zhang2019interpreting,frosst2017distilling,tan2018distill,liu2018improving}.
The work in~\cite{achille2018emergence} quantifies the disentanglement of representations based on information theory.
The key idea of disentanglement-based interpretability is to align the learned representations' distribution to a standard Gaussian distribution, and then assign concepts to each independent dimension of the representations.
Instead of learning disentangled representations from scratch, we focus on disentangling the entire architecture of a pre-trained DNN to interpret its inner workings.
To this end, an objective function is constructed to constrain the information between successive layers of DNNs for the architecture disentanglement.

% \textbf{Dynamic Routings of DNNs.} Dynamic routings focus on studying how to select a routing for each input during inference~\cite{bengio2015conditional, bengio2013estimating, bolukbasi2017adaptive}.
% %
% For instance, Distillation Guided Routing~\cite{wang2018interpret} uses dynamic routings to interpret the neural networks, which finds paths for each \textit{individual input}, and interprets DNNs by comparing \textit{intra-class} samples.
% %
% In contrast, our work disentangles the network architecture into paths with different \textit{semantic concepts}, and interprets DNNs according to the semantic difference between \textit{inter-class} concepts.
% 

\section{Method}
\label{Method}
\subsection{Problem Formulation}
NAD aims to decompose a pre-trained DNN into a set of sub-architectures for different tasks.
As disentangling the decision-making process in the classifier leads to fixed one-hot vectors in the output layer, we focus on the information flows in the feature extractor of one DNN.
Let $c$ denote the class of the target classification task to be disentangled from the DNNs.
The image $\boldsymbol{x}^c$ is sampled from the set of images $\boldsymbol{X}^c$ labeled with $\boldsymbol{y}^c$.
We define the layers in the feature extractor as $n$ functions $f_0(\cdot), f_1(\cdot), ..., f_{n-1}(\cdot)$ from bottom to top, where $f_{n}(\cdot)$ is the classifier.
For the input $\boldsymbol{x}^c$, we can obtain the feature maps in the original network as:
\begin{equation}
\begin{split}
\label{eq1}
\boldsymbol{r}_0^c, \boldsymbol{r}_1^c, ..., \boldsymbol{r}_{n-1}^c, \widetilde{\boldsymbol{y}}^c=f_0(\boldsymbol{x}^c), &f_1\big(f_0(\boldsymbol{x}^c)\big),\\
...,&f_n\big(f_{n-1}(...f_0(\boldsymbol{x}^c)...)\big),
\end{split}
\end{equation}
where $\boldsymbol{r}_0^c, \boldsymbol{r}_1^c, ...,\boldsymbol{r}_{n-1}^c$ denote the output feature maps in the feature extractor, and $\widetilde{\boldsymbol{y}}^c$ denotes the output of the classifier.
As there is a one-to-one relationship between the feature maps and the filters of the convolutional layers, selecting the feature maps of each layer is equivalent to selecting the filters for constructing the sub-architectures.
Therefore, NAD finds the minimal combinations of feature maps in the feature extractor, so that the classifier has the maximal possibility to predict the label $\boldsymbol{y}^c$.
The sub-architecture for the task of classifying $c$ is constructed using the selected filters.
\subsection{Representation Disentanglement}
The target of NAD is similar to that of the representation disentanglement with the information bottleneck theory~\cite{tishby2000information}.
Given the input $\boldsymbol{x}^c\in \boldsymbol{X}^c$ with its representations $\boldsymbol{r}_{n-1}^c\in \boldsymbol{R}_{n-1}^c$ and label $\boldsymbol{y}^c\in \boldsymbol{Y}^c$, the objective function of the information bottleneck is defined as:
\begin{equation}
\begin{split}
\label{eq2}
\mathcal{L}_{IB}=\beta\cdot\mathcal{I}(\boldsymbol{x}^c; \boldsymbol{r}_{n-1}^c) - \mathcal{I}(\boldsymbol{r}_{n-1}^c; \boldsymbol{y}^c),
\end{split}
\end{equation}
where $\mathcal{I}(\cdot; \cdot)$ denotes the mutual information shared between two random variables, and $\beta>0$ is a hyperparameter.
The first term of the objective function indicates that the representations should use the input information as little as possible, while the second term ensures that they also keep as much information as possible for accomplishing downstream tasks.
After derivation,\footnote{Due to space limitations, we only show the final formulations in the main body of this paper. The derivations can be found in Appendix.} the variational upper bound of Eq.~\ref{eq2} is:
\begin{equation}
\begin{split}
\label{eq3} % \Big] \mathbb{E}_{x^c\sim P(x^c)}\Big[
\mathcal{\widetilde{L}}_{IB}=&\mathbb{E}_{\boldsymbol{x}^c\sim P(\boldsymbol{x}^c)}\Big[\beta\cdot KL\big[P(\boldsymbol{r}_{n-1}^c|\boldsymbol{x}^c)||Q(\boldsymbol{r}_{n-1}^c)\big] \\
-&\mathbb{E}_{\boldsymbol{r}_{n-1}^c\sim P(\boldsymbol{r}_{n-1}^c|\boldsymbol{x}^c)}\big[\log Q(\boldsymbol{y}^c|\boldsymbol{r}_{n-1}^c)\big]\Big],
\end{split}
\end{equation}
where $KL$ denotes the Kullback-Leibler (KL) divergence between two distributions, and $Q(\cdot)$ is the predefined distribution for variational approximation.
Eq.~\ref{eq3} has the same formulation as the objective function in the standard variational autoencoder (VAE)~\cite{kingma2013auto}.
By defining $Q(\cdot)$ as the Gaussian distribution and enlarging $\beta$ for regularization, $\beta$-VAE~\cite{higgins2017beta} disentangles the factors into every single dimension of the representations under a self-supervised setting.
\subsection{Architecture Disentanglement}
For the architecture disentanglement, we modify the representation disentanglement from a single constraint to multiple constraints, and disentangle the factors between successive layers.
Inspired by~\cite{dai2018compressing,tishby2015deep}, the networks can be interpreted as a Markov chain between successive representations:
\begin{equation}
\begin{split}
\label{eq4}
\boldsymbol{y}^c\rightarrow \boldsymbol{x}^c \rightarrow \boldsymbol{r}_0^c \rightarrow \boldsymbol{r}_1^c \rightarrow ... \rightarrow \boldsymbol{r}_{n-1}^c \rightarrow \widetilde{\boldsymbol{y}}^c,
\end{split}
\end{equation}
where the trained network extracts the representations for predicting the labels.
The goal is to keep the information for classifying class $c$, while removing the redundant information for other classes.
Therefore, between each adjacent layer, we reward the information related to the classification of class $c$, while penalizing the information of other classes.

For the $i$-th hidden layer, \ie, when $1\le i < n$, the objective function can be written as:
\begin{equation}
\begin{split}
\label{eq5}
\mathcal{L}^c_i= \beta\cdot\mathcal{I}(\boldsymbol{r}^c_{i-1};\widetilde{\boldsymbol{r}}^c_i) - \mathcal{I}(\boldsymbol{r}^c_i;\widetilde{\boldsymbol{r}}^c_i),
\end{split}
\end{equation}
where $\widetilde{\boldsymbol{r}}^c_i$ denotes the representation after constraining the information.
As with Eq.~\ref{eq3}, the variational upper bound of Eq.~\ref{eq5} can be derived as:
\begin{equation}
\begin{split}
\label{eq6} %\Big] \mathbb{E}_{r_{i-1}^c\sim P(r_{i-1}^c)}\Big[
\mathcal{\widetilde{L}}^c_i=\mathbb{E}_{\boldsymbol{r}_{i-1}^c\sim P(\boldsymbol{r}_{i-1}^c)}\Big[\beta\cdot KL\big[P(\widetilde{\boldsymbol{r}}_i^c|\boldsymbol{r}_{i-1}^c)||Q(\widetilde{\boldsymbol{r}}_i^c)\big]& \\
- \mathbb{E}_{\boldsymbol{r}_i^c\sim P(\boldsymbol{r}_i^c|\boldsymbol{r}_{i-1}^c)}\big[\log Q(\widetilde{\boldsymbol{r}}_i^c|\boldsymbol{r}_i^c)\big]&\Big].
\end{split}
\end{equation}
To optimize Eq.~\ref{eq6}, we specify the parametric forms of the distributions, which is usually done by assuming a Gaussian distribution for regression and a multinomial distribution for classification.
Therefore, we assume $Q(\widetilde{\boldsymbol{r}}_i^c), Q(\widetilde{\boldsymbol{r}}_i^c|\boldsymbol{r}_i^c)$ as the standard Gaussian distribution for regressing the representations, and define $\widetilde{\boldsymbol{r}}_i^c$ as:
\begin{equation}
\begin{split}
\label{eq7}
\widetilde{\boldsymbol{r}}_i^c&=(\boldsymbol{\mu}_i^c+\boldsymbol{\epsilon}_i\cdot\boldsymbol{\sigma}^c_i)\cdot \boldsymbol{r}_i^c\\
&=(\boldsymbol{\mu}^c_i+\boldsymbol{\epsilon}_i\cdot\boldsymbol{\sigma}^c_i)\cdot f_i(\boldsymbol{r}_{i-1}^c),
\end{split}
\end{equation}
where $\boldsymbol{\epsilon}_i$ is random noise sampled from the standard Gaussian distribution $\mathcal{N}(\boldsymbol{0}, \boldsymbol{I})$, and $\boldsymbol{\mu}^c_i, \boldsymbol{\sigma}^c_i$ are the learnable parameters.
With Eq.~\ref{eq7}, we can correlate the information in successive layers as well as expanding the KL divergence term in Eq.~\ref{eq6}, which defines the conditional probability as:
\begin{equation}
\begin{split}
\label{eq8}
P(\widetilde{\boldsymbol{r}}_i^c|\boldsymbol{r}_{i-1}^c)&=\mathcal{N}\Big(f_i(\boldsymbol{r}_{i-1}^c)\cdot \boldsymbol{\mu}^c_i,diag\big[(f_i(\boldsymbol{r}_{i-1}^c)\cdot \boldsymbol{\sigma}^c_i)^2\big]\Big)\\
&=\mathcal{N}\Big(\boldsymbol{r}_i^c\cdot \boldsymbol{\mu}^c_i, diag\big[(\boldsymbol{r}_i^c\cdot \boldsymbol{\sigma}^c_i)^2\big]\Big).
\end{split}
\end{equation}
Combining the parametric forms of the distributions $Q(\widetilde{\boldsymbol{r}}_i^c), Q(\widetilde{\boldsymbol{r}}_i^c|\boldsymbol{r}_i^c)$, and $P(\widetilde{\boldsymbol{r}}_i^c|\boldsymbol{r}_{i-1}^c)$, the constraint in Eq.~\ref{eq6} for the $i$-th hidden layer can be derived as:
\begin{equation}
\begin{split}
\label{eq9}
\mathcal{\widetilde{L}}^c_i&=\frac{\beta}{2}\big((\boldsymbol{r}_i^c\cdot \boldsymbol{\sigma}^c_i)^2-\log{(\boldsymbol{r}_i^c\cdot \boldsymbol{\sigma}^c_i)^2}+(\boldsymbol{r}_i^c\cdot \boldsymbol{\mu}^c_i)^2-1\big)\\
&+\frac{1}{2}\big(||\boldsymbol{r}^c_i - \boldsymbol{\mu}^c_i\cdot \boldsymbol{r}^c_i||_2^2 + \log 2\pi + \log{(\boldsymbol{r}_i^c\cdot \boldsymbol{\sigma}^c_i)^2}\big).
\end{split}
\end{equation}

For the classifier's output, \ie, when $i=n$, we maximize the information between $\widetilde{\boldsymbol{y}}^c$ and $\boldsymbol{y}^c$ to guarantee that the constrained representations will keep the information for class $c$.
The objective function is:
\begin{equation}
\begin{split}
\label{eq10}
\mathcal{L}^c_n= - \mathcal{I}(\widetilde{\boldsymbol{y}}^c;\boldsymbol{y}^c).
\end{split}
\end{equation}
Correspondingly, the variational upper bound of Eq.~\ref{eq10} is:
\begin{equation}
\begin{split}
\label{eq11}
\mathcal{\widetilde{L}}^c_n=\mathbb{E}_{\boldsymbol{r}_{n-1}^c\sim P(\boldsymbol{r}_{n-1}^c)}\Big[\mathbb{E}_{\widetilde{\boldsymbol{y}}^c\sim P(\widetilde{\boldsymbol{y}}^c|\boldsymbol{r}_{n-1}^c)}\big[-\log Q(\boldsymbol{y}^c|\widetilde{\boldsymbol{y}}^c)\big]\Big].
\end{split}
\end{equation}
By assuming the variational distribution $Q(\boldsymbol{y}^c|\widetilde{\boldsymbol{y}}^c)$ as the multinomial distribution for classifying $\boldsymbol{y}^c$, we obtain the cross entropy loss:
\begin{equation}
\begin{split}
\label{eq12}
\mathcal{\widetilde{L}}^c_n=&-\boldsymbol{y}^c\log f_n(\boldsymbol{r}^c_{n-1})\\
&-(1-\boldsymbol{y}^c)\log\big(1-f_n(\boldsymbol{r}^c_{n-1})\big)\\
=&-\boldsymbol{y}^c\log\widetilde{\boldsymbol{y}}^c-(1-\boldsymbol{y}^c)\log(1-\widetilde{\boldsymbol{y}}^c).
\end{split}
\end{equation}
\subsection{Optimization of Objective Function}
To simplify the optimization, we reduce the noise level by fixing $\boldsymbol{\epsilon}_i$ in Eq.~\ref{eq7} to its mean value, \ie, $\boldsymbol{0}$, which frees $\boldsymbol{\sigma}^c_i$ to be any vector that does not affect the optimization process.
Then, Eq.~\ref{eq9} becomes a much easier formulation to optimize:
\begin{equation}
\begin{split}
\label{eq13}
\mathcal{\widetilde{L}}^c_i=\beta\cdot(\boldsymbol{r}_i^c\cdot \boldsymbol{\mu}^c_i)^2
+||\boldsymbol{r}^c_i - \boldsymbol{\mu}^c_i\cdot \boldsymbol{r}^c_i||_2^2,
\end{split}
\end{equation}
where the first term regularizes $\boldsymbol{\mu}_i$, and the second term constrains the reconstruction error for $\boldsymbol{r}_i^c$.
Combining with Eq.~\ref{eq12} and Eq.~\ref{eq13}, the overall objective function is:
\begin{equation}
\begin{split}
\label{eq14}
\mathcal{\widetilde{L}}^c=\mathbb{E}_{\boldsymbol{x}^c\in \boldsymbol{X}^c}\Big[\sum_{i=0}^{n}\widetilde{L}_i^c\Big].
\end{split}
\end{equation}
To constrain the values to $(0,1)$, we activate $\boldsymbol{\mu}_i^c$ using the Sigmoid function.
Stochastic gradient descent is adopted to update $\boldsymbol{\mu}_i^c$, and L2 normalization is applied to the gradient for fast convergence.
\begin{algorithm}[tb]
 \caption{Neural Architecture Disentanglement}
 \label{alg1}
\begin{algorithmic}
 \STATE {\bfseries Input:} The pre-trained network to be disentangled, the data $\boldsymbol{X}^c$ labeled with class $c$, the parameter $\boldsymbol{\mu}^c$ to be learned, the learning rate $\alpha$, and the hyperparameter $\beta$.
 \STATE {\bfseries Output:} The learned $\boldsymbol{\mu}^c$ for class $c$.
 \STATE Initialize $\boldsymbol{\mu}^c$ randomly.
 \REPEAT
 \STATE Sample $\boldsymbol{x}^c$ from $\boldsymbol{X}^c$.
 \STATE For $i=1,...,n-1$, calculate $\widetilde{\boldsymbol{y}}^c$ and $\boldsymbol{r}_i^c$ by Eq.~\ref{eq1}.
 \STATE For $i=1,...,n-1$, calculate $\widetilde{\boldsymbol{r}}^c_i$ by Eq.~\ref{eq7}.
 \STATE Calculate the loss by Eq.~\ref{eq14} with Eq.~\ref{eq12} and Eq.~\ref{eq13}.
 \STATE For $i=1,...,n-1$, update $\boldsymbol{\mu}_i^c$ by Eq.~\ref{eq15}.
 \UNTIL{Convergence}
 \STATE For $i=1,...,n-1$, discretize $\boldsymbol{\mu}_i^c$ by Eq.~\ref{eq16}.
\end{algorithmic}
\end{algorithm}
The update of $\boldsymbol{\mu}^c_i$ is defined as:
\begin{equation}
\begin{split}
\label{eq15}
{\boldsymbol{\mu}_i^c}^{'}=\boldsymbol{\mu}_i^c-\alpha\cdot norm(\frac{\partial \mathcal{\widetilde{L}}^c}{\partial \boldsymbol{\mu}_i^c}),
\end{split}
\end{equation}
where $\alpha$ is the learning rate.
The pre-trained networks are fixed during optimization.
After convergence, we discretize the continuous $\boldsymbol{\mu}_i^c$ to binary values to form the disentangled sub-architecture for classifying classifier $c$:
\begin{equation}
\begin{split}
\label{eq16}
\boldsymbol{\mu}^c_i = 
 \begin{cases}
 1, &\text{if $\boldsymbol{\mu}^c_i>0.5$}\\
 0, &\text{if $\boldsymbol{\mu}^c_i\le0.5$},
 \end{cases}
\end{split}
\end{equation}
where $0.5$ is the threshold from the Sigmoid function.
After obtaining $\boldsymbol{\mu}^c_i$ with $i=0,..,n-1$ for all classes $c\in C$, we have naturally disentangled the original network.
Alg.~\ref{alg1} summarizes the overall procedure of NAD.

\section{Experiments}
\label{Experiments}
In this section, we investigate whether, where, and how NAD occurs to understand the inner workings of DNNs.
We first check the classification results to ensure the information of classifying specific classes is disentangled with the sub-architectures.
Then, two indexes measuring the similarity and class hit rate of sub-architectures are designed to study where the networks begin to disentangle.
Together with the visualizations of the activated feature maps, we try to understand how the entire networks work.
In the end, we provide an explanation for why DNNs sometimes give wrong predictions.
\subsection{Experimental Settings}
\noindent\textbf{Datasets and Network Architectures.} We conduct experiments on both object-based and scene-based datasets, \ie, ImageNet~\cite{deng2009imagenet} and Place365~\cite{zhou2017places}.
These datasets are organized in terms of the WordNet hierarchy~\cite{miller1998wordnet}, with each node being depicted by hundreds of images.
We disentangle the original architectures using the training set, and the validation set is used to study the properties of DNNs.
\begin{figure}[t]
 \centering
 \begin{subfigure}{1\linewidth}
 \centering
 \caption{Classification results on ImageNet} \vspace{-2mm}
 \includegraphics[width=0.99\linewidth]{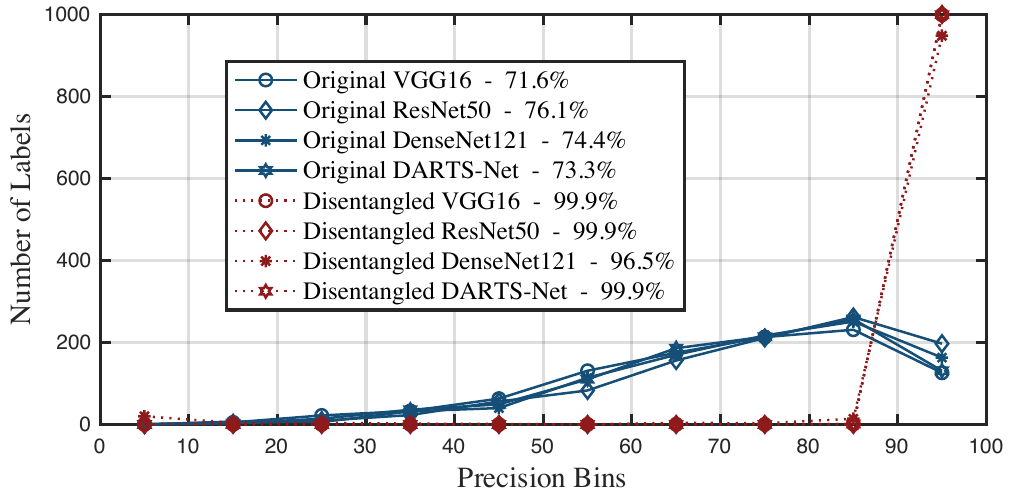}
 \label{fig2-1}
 \end{subfigure}
 \begin{subfigure}{1\linewidth}
 \centering
 \vspace{0.2cm}
 \caption{Classification results on Place365} \vspace{-2mm}
 \includegraphics[width=0.99\linewidth]{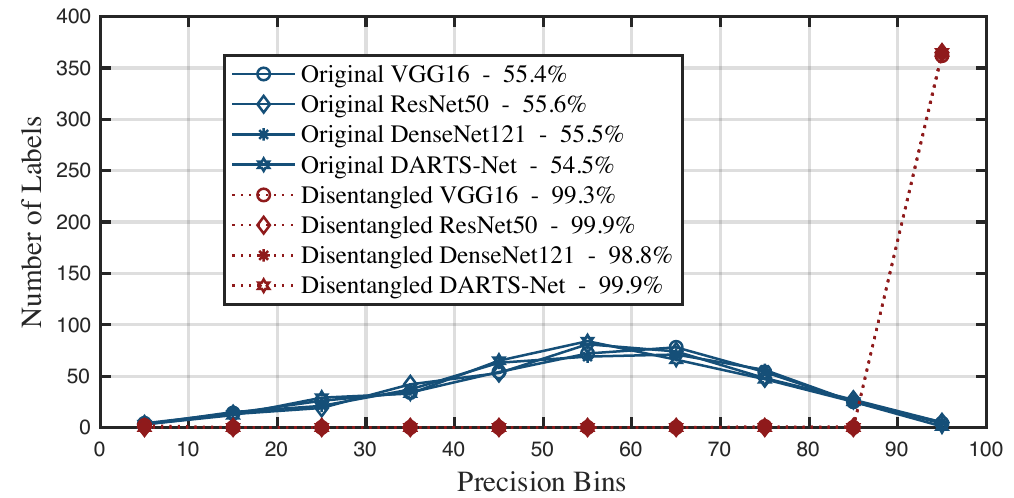} 
 \label{fig2-2}
 \end{subfigure}
 \caption{
 Histograms of the Top@1 classification accuracies (\%) of the original architectures and the disentangled sub-architectures.
 The horizontal axis denotes the bins of Top@1 classification precision for one single label, in which the values are discretized into ten intervals, \ie, $[0,10],(10,20],...,(90,100]$.
 The vertical axis denotes the number of classes whose classification accuracies fall into the same discretized bins.
 The solid lines show the results of the original architectures, and the dotted lines show the results of the disentangled sub-architectures.
 The average top@1 classification accuracies are listed in the legends.
 % to represent the individual concepts
 The classification results indicate that the disentangled sub-architectures are linked with their corresponding classes.
 }%\vspace{-3mm}
 \label{fig2}
\end{figure}
We select four network architectures, \ie, VGG16~\cite{simonyan2014very}, ResNet50~\cite{he2016deep}, DenseNet121~\cite{huang2017densely} and DARTS-Net~\cite{liu2018darts}, to conduct our experiments.
The models are pre-trained on the above two datasets, and their parameters are fixed when disentangling the architectures.
The connection types include the direct connection in VGG, the plain skip-connection in ResNet, the dense skip-connection in DenseNet, and the automatically-searched skip-connection in DARTS-Net.
The architecture of DARTS-Net is searched on the CIFAR10~\cite{krizhevsky2009learning} dataset.
%with Pytorch~\cite{paszke2019pytorch}
%

\noindent\textbf{Implementation Details.} We set the learning rate $\alpha=0.1$ and the iterations $N=20$ to disentangle the DNNs for each class.
We use the Elbow method to determine the hyperparameter $\beta$ for balancing the regularization term and the reconstruction term in the final objective function.\footnote{More detailed results can be found in Appendix.}
Following the experimental settings in~\cite{selvaraju2017grad,zhou2016learning}, feature map visualization is performed by bilinearly interpolating the size of feature maps to the size of input images and visualizing the top $5\%$ of activated pixels.
\subsection{Properties of NAD}
\subsubsection{Does Disentanglement Occur?}
%original general-purpose architectures and the disentangled special-purpose
We check the single-class classification results for the disentangled sub-architectures and the original architectures, to investigate whether the disentanglement occurs in DNNs.
Specifically, we perform classification with images from the same classes, by inputting them into their corresponding sub-architectures and the original architectures.
The Top@1 classification accuracies (\%) are recorded for each class, and the values are discretized into bins of $[0,10], (10,20],...,(90,100]$.
We accumulate the number of classes whose accuracies fall into the same bins.
From Fig.~\ref{fig2}, we can see that the accuracies increase after disentanglement.
For example, on ImageNet, the distributions of the original architectures mainly fall into the interval of $(70,90]$, while those of the disentangled sub-architectures are squashed into the interval of $(90,100]$.
These results suggest that the sub-architectures can be disentangled to precisely relate to the assigned classes.
\begin{figure}[t]
 \centering
 \includegraphics[width=2.9in]{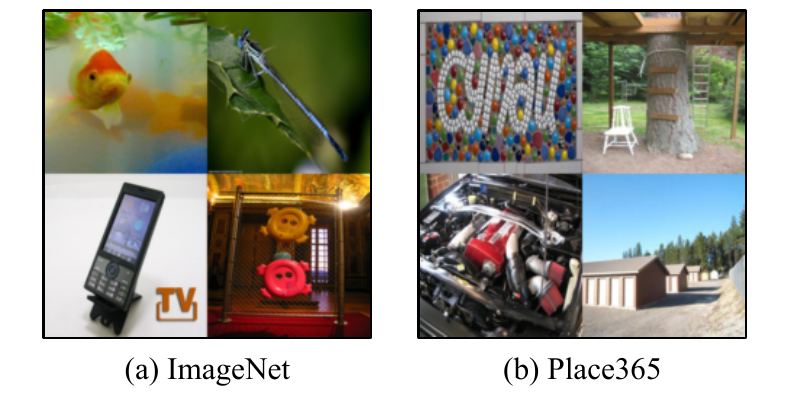}
 \caption{
 Examples of randomly combined images for calculating the hit rate of the disentangled sub-architectures.
 Example (a) combines the images with `label-concept': `1-Goldfish', `320-Damselfly', `487-MobilePhone', and `489-ChainLinkFence' from the validation set of ImageNet.
 Example (b) combines the images with `label-concept': `34-BallPit', `339-TreeHouse', `28-AutoFactory', and `221-ManufacturedHome' from the validation set of Place365.
 }%\vspace{-3mm}
 \label{fig3}
\end{figure}
\begin{figure*}[ht]
 \centering
\begin{subfigure}{0.495\textwidth}
 \centering
 \caption{Results of VGG16} \vspace{-2mm}
 \includegraphics[width=0.99\linewidth]{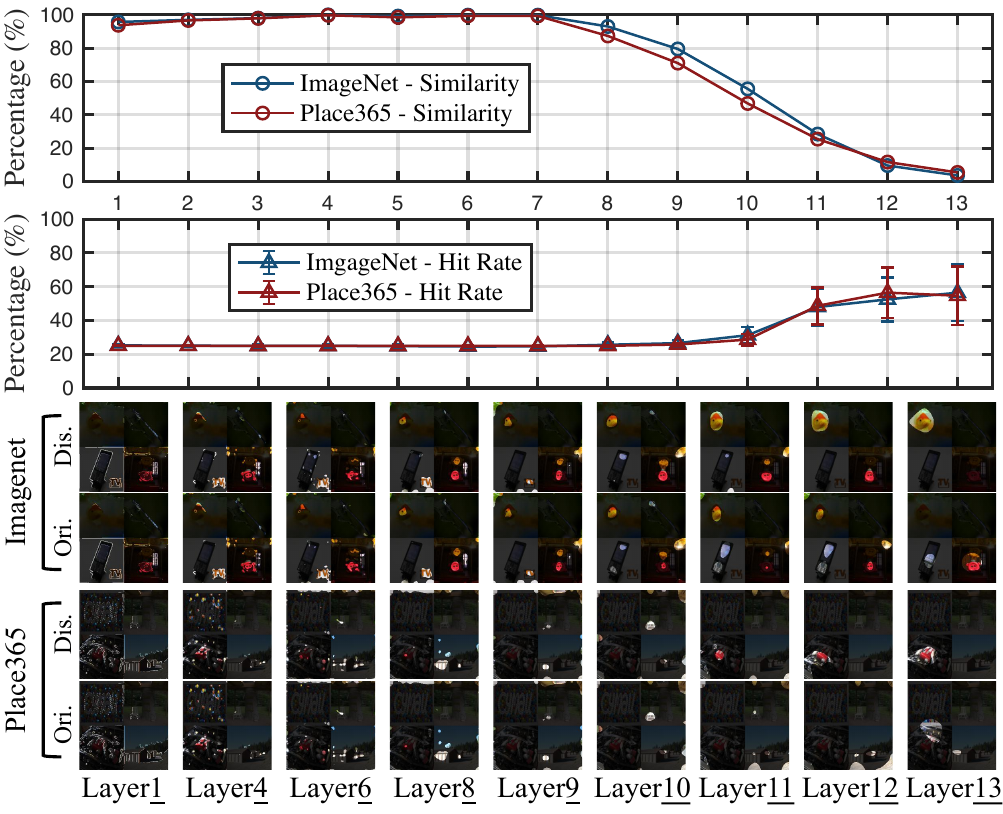}
 \label{fig4-1}
\end{subfigure}
\vspace{0.2cm}
\begin{subfigure}{0.495\textwidth}
 \centering
 \caption{Results of ResNet50} \vspace{-2mm}
 \includegraphics[width=0.99\linewidth]{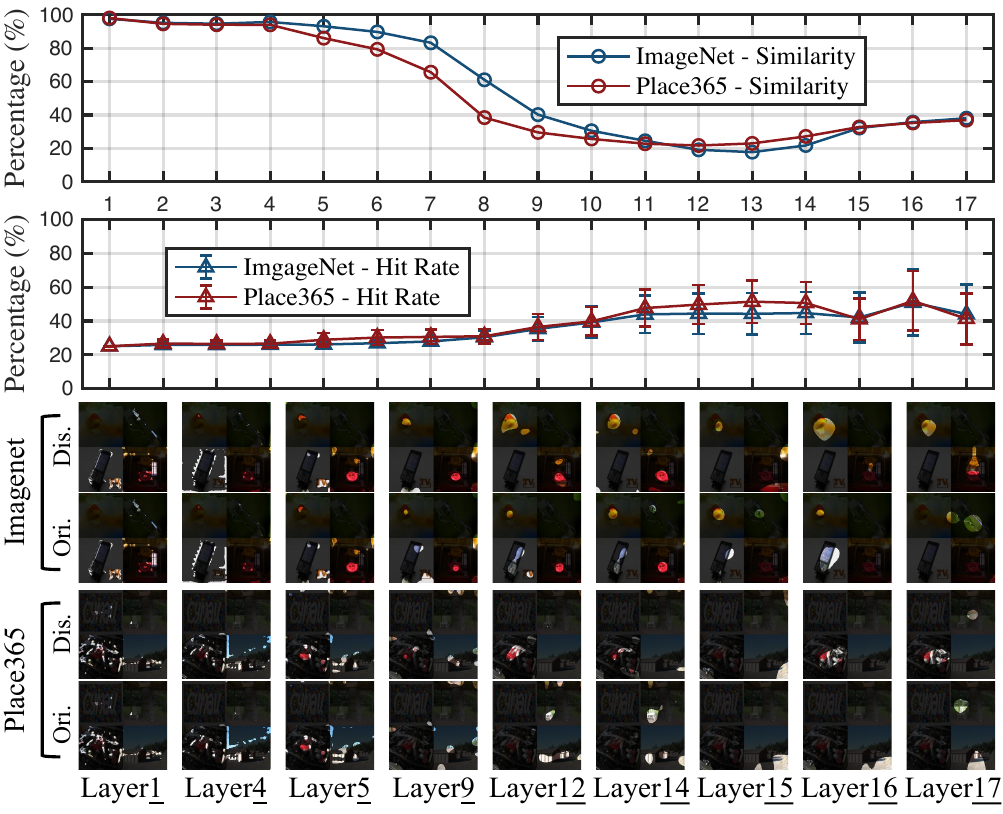}
 \label{fig4-2}
\end{subfigure}
\begin{subfigure}{0.495\textwidth}
 \centering
 \caption{Results of DenseNet121} \vspace{-2mm}
 \includegraphics[width=0.99\linewidth]{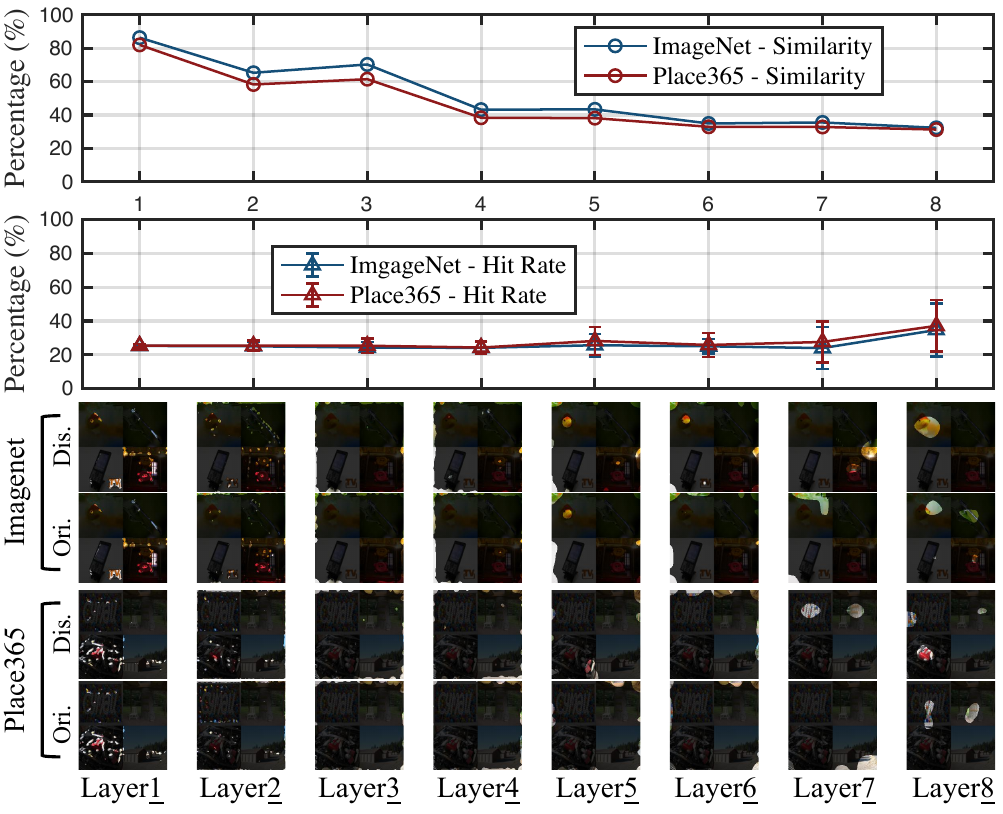} 
 \label{fig4-3}
\end{subfigure}
\begin{subfigure}{0.495\textwidth}
 \centering
 \caption{Results of DARTS-Net} \vspace{-2mm}
 \includegraphics[width=0.99\linewidth]{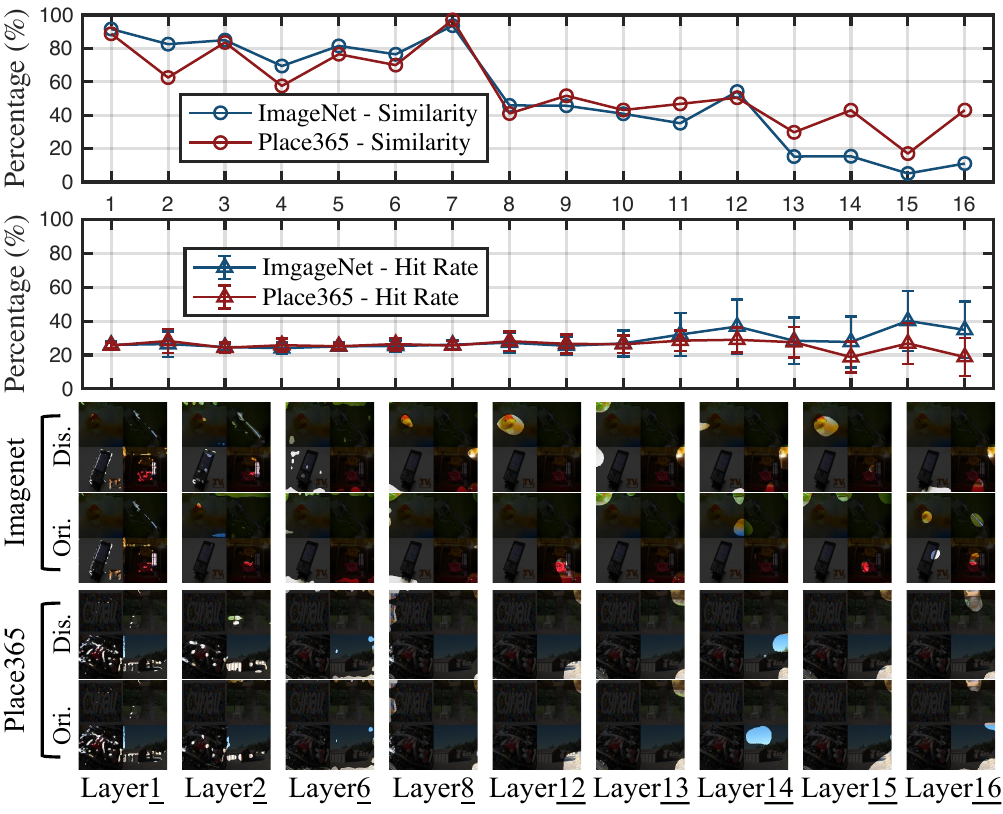} 
 \label{fig4-4}
\end{subfigure}
\caption{Results of the similarity, hit rate and feature map visualization. Best viewed in color, zoomed in for details.}%\vspace{-3mm}
\label{fig4}
\end{figure*}
\vspace{-0.2cm}
\subsubsection{Where Does the Disentanglement Occur?}
Considering that the sub-architectures can output the target classes regardless of which images are given, we design hit rate experiments to check if the disentangled sub-architectures can distinguish the corresponding classes from the randomly-combined images.
%
% After obtaining the disentangled sub-architectures, we investigate where the disentanglement occurs throughout the networks.
%
To do so, we introduce two indexes, \ie, the similarity and the hit rate for the sub-architectures.

\noindent\textbf{Similarity Between Sub-Architectures.}
The first index is the similarity measurement.
As the sub-architectures can be regarded as sets combined with selected filters, the similarity can be calculated by the Jaccard coefficient, which is defined as the intersection size divided by the union size between two sets.
We compute the average similarity for each layer between all pairs of disentangled sub-architectures.
The results are shown in the first rows of Figs.~\ref{fig4-1},~\ref{fig4-2},~\ref{fig4-3} and~\ref{fig4-4}.
%, which verifies that the bottom layers extract low-level information such as edge and color, while the top layers extract high-level semantics such as parts of objects
Overall, the similarity between bottom layers is higher than that of top layers.
Correspondingly, classes share low-level information in the bottom layers, and high-level semantics are gradually combined in the middle layers for classification.
Therefore, the disentanglement tends to start in the middle layers, and the concrete concepts will gradually emerge from the middle layers to the top layers.

For VGG16 in Fig.~\ref{fig4-1}, the similarity is roughly above 90\% in the first seven layers, which indicates that the low-level information extraction is conducted in these layers.
Meanwhile, in the architectures with skip-connections, the results become more complex to analyze.
For ResNet50 in Fig.~\ref{fig4-2}, the similarity begins to decrease early in the 4th layer and reaches the lowest point between the 11th and 14th layer.
For DenseNet121 in Fig.~\ref{fig4-3}, the similarity is low in all layers.
For DARTS-Net in Fig.~\ref{fig4-4}, the similarity values change drastically, especially in the 8th, 12th, and 15th layers.
We believe that these results arise from the information fusion caused by the skip-connections.
The high-level semantic information extracted in the top layers is brought forward to the middle or bottom layers by the skip-connection.
To further verify our statement, we design the second index.
\begin{table}[t]
\centering
 \scalebox{0.82}[0.82]{
 \centering
\begin{tabular}{cc|c|c|c|c}
\toprule
\textbf{} & & \multicolumn{1}{c|}{\textbf{VGG16}} & \multicolumn{1}{c|}{\textbf{ResNet50}} & \multicolumn{1}{c|}{\textbf{DenseNet121}} & \textbf{DARTS-Net} \\ \hline
\multicolumn{1}{c|}{\textbf{}} & Mis. & 78.59 & 59.38 & 53.35 & 60.02 \\
\multicolumn{1}{c|}{\textbf{Img.}} & Oth. & 73.74 & 57.09 & 51.41 & 52.97 \\ \cline{2-6} 
\multicolumn{1}{c|}{} & \textit{Dif.} & \textbf{-4.85} & \textbf{-2.29} & \textbf{-1.94} & \textbf{-7.05} \\ \hline \hline
\multicolumn{1}{c|}{} & Mis. & 81.87 & 65.49 & 43.95 & 61.96 \\
\multicolumn{1}{c|}{\textbf{Plc.}} & Oth. & 78.75 & 60.34 & 42.02 & 55.97 \\ \cline{2-6} 
\multicolumn{1}{c|}{} & \textit{Dif.} & \textbf{-3.12} & \textbf{-5.15} & \textbf{-1.93} & \textbf{-5.99} \\
\bottomrule
\end{tabular}}
\caption{
Average similarity of two different pairs of sub-architectures.
`Mis.' denotes the average similarity between the sub-architectures of the correct labels and the sub-architectures of the misclassified labels.
`Oth.' denotes the average similarity between the sub-architectures of the correct labels and the sub-architectures of all other labels except the misclassified ones.
`Dif.' denotes the difference between `Mis.' and `Oth.'.
`Img.' denotes the ImageNet dataset, and `Plc.' denotes Place365.
All the values of `Dif.' are greater than zero, which indicates that the misclassified input images have a high probability of being assigned to classes with similar sub-architectures to the correct ones.
}%\vspace{-3mm}
\label{tab1}
\end{table}

\noindent\textbf{Hit Rate of Sub-Architectures.}
For the hit rate, we first combine four images with randomly selected labels to obtain the test images.
Fig.~\ref{fig3} shows two examples of the combined images.
We input the combined images to the disentangled sub-architectures and visualize the activated feature maps.
We accumulate the number of correct hits at each pixel of the feature maps, and divide them by all the activated pixels.
A correct hit means the activated pixel is located on the image of the correct class in the combined image.
% by randomly combine four images
We carry out this experiment $1,000$ times using the randomly selected classes, and the mean values are computed for each layer.
Note that using the original architectures in this experiment makes no sense, as the activated feature maps for the four selected classes in the combined images are the same.
The hit rate results are shown in the second rows of Figs.~\ref{fig4-1},~\ref{fig4-2},~\ref{fig4-3} and~\ref{fig4-4}.
The third and fourth rows show examples of the visualized feature maps for the ImageNet and Place365 datasets,\footnote{More results can be found in Appendix.} respectively.
`Dis.' denotes the feature maps from the disentangled sub-architectures for the class `Goldfish' in ImageNet and `AutoFactory' in Place365.
`Ori.' denotes the visualized feature maps of the original architectures.
\begin{figure*}[ht]
 \centering
 \begin{subfigure}{0.24\textwidth}
 \centering
 \caption{VGG16 on ImageNet} \vspace{-2mm}
 \includegraphics[width=1.62in]{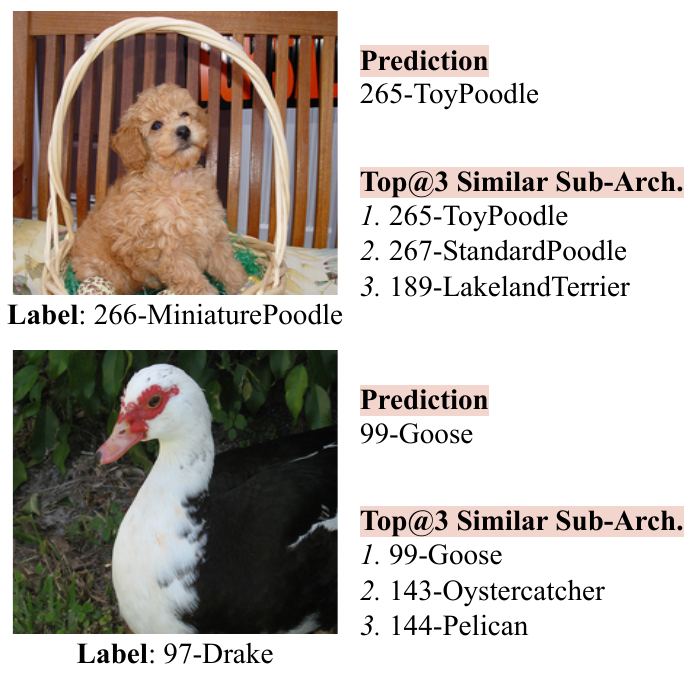}
 \label{fig5-1}
 \end{subfigure}
 \begin{subfigure}{0.24\textwidth}
 \centering
 \caption{ResNet50 on ImageNet} \vspace{-2mm}
 \includegraphics[width=1.62in]{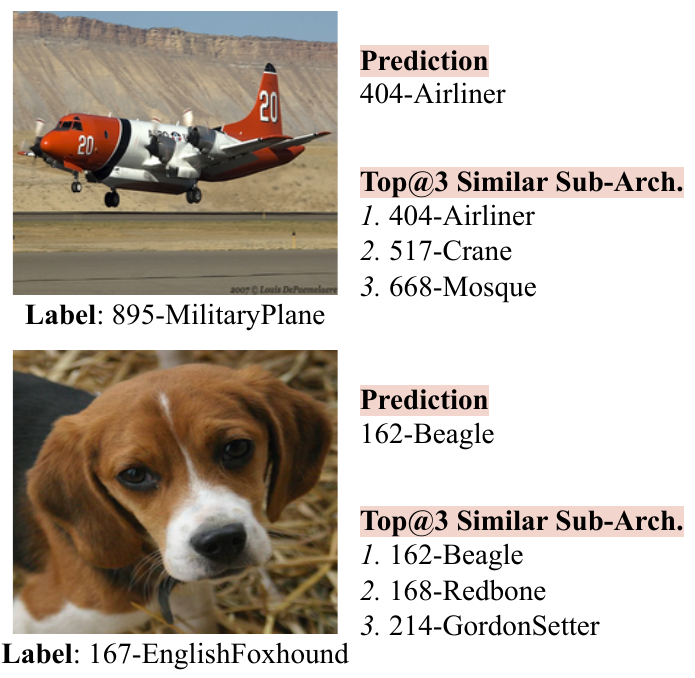}
 \label{fig5-2}
 \end{subfigure}
 \begin{subfigure}{0.24\textwidth}
 \centering
 \caption{DenseNet121 on ImageNet} \vspace{-2mm}
 \includegraphics[width=1.62in]{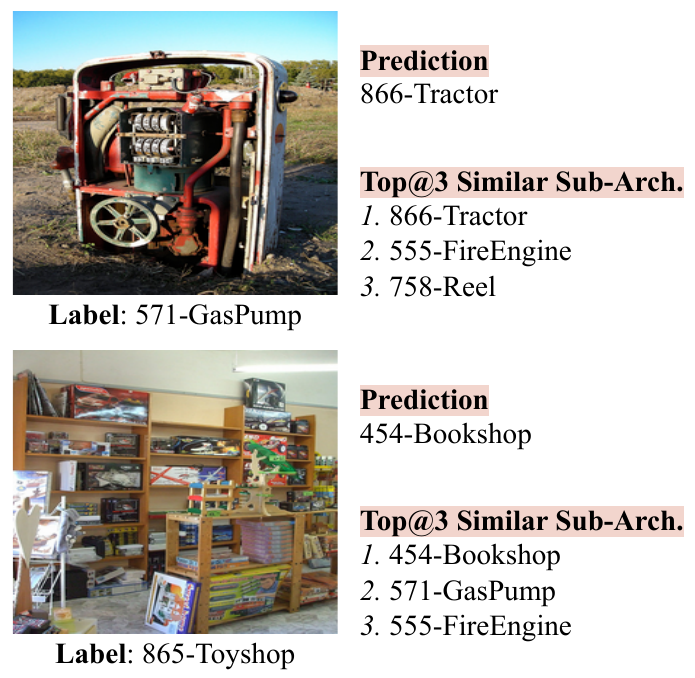}
 \label{fig5-3}
 \end{subfigure}
 \begin{subfigure}{0.24\textwidth}
 \centering
 \caption{DARTS-Net on ImageNet} \vspace{-2mm}
 \includegraphics[width=1.62in]{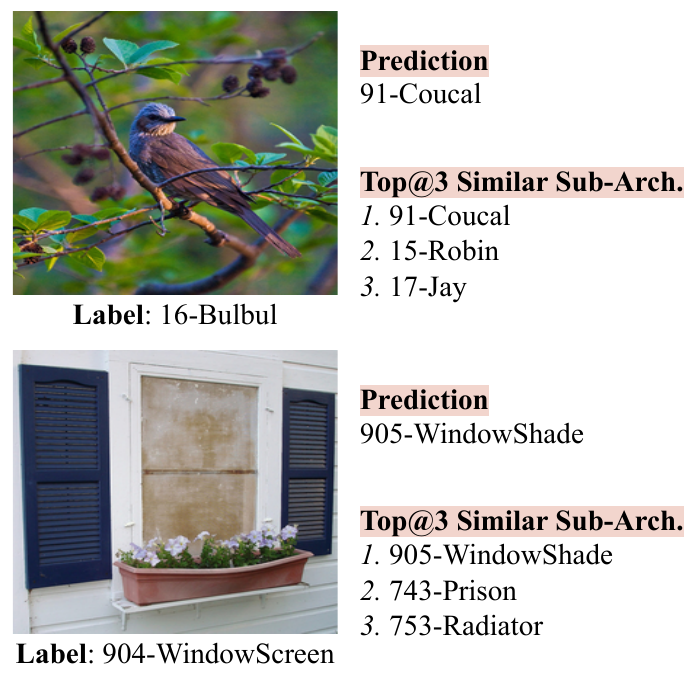}
 \label{fig5-4}
 \end{subfigure}
 \begin{subfigure}{0.24\textwidth}
 \centering
 \caption{VGG16 on Place365} \vspace{-2mm}
 \includegraphics[width=1.62in]{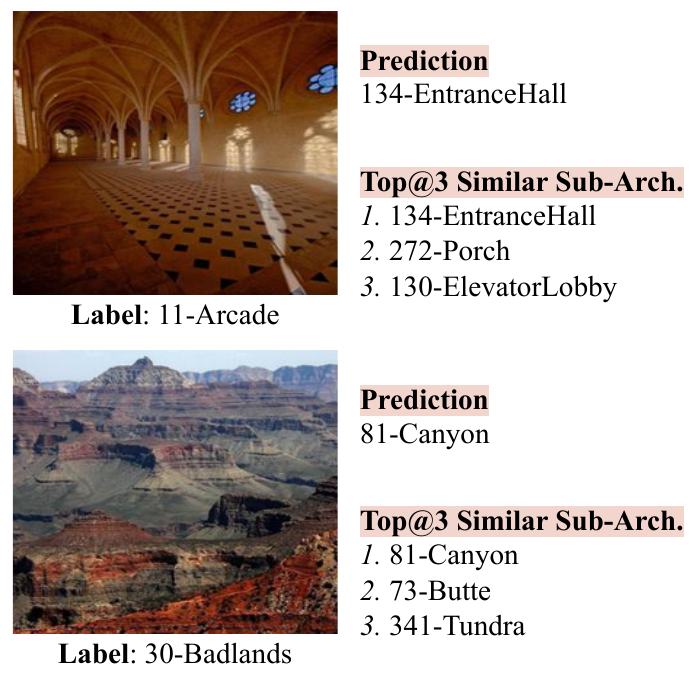}
 \label{fig5-5}
 \end{subfigure}
 \begin{subfigure}{0.24\textwidth}
 \centering
 \caption{ResNet50 on Place365} \vspace{-2mm}
 \includegraphics[width=1.62in]{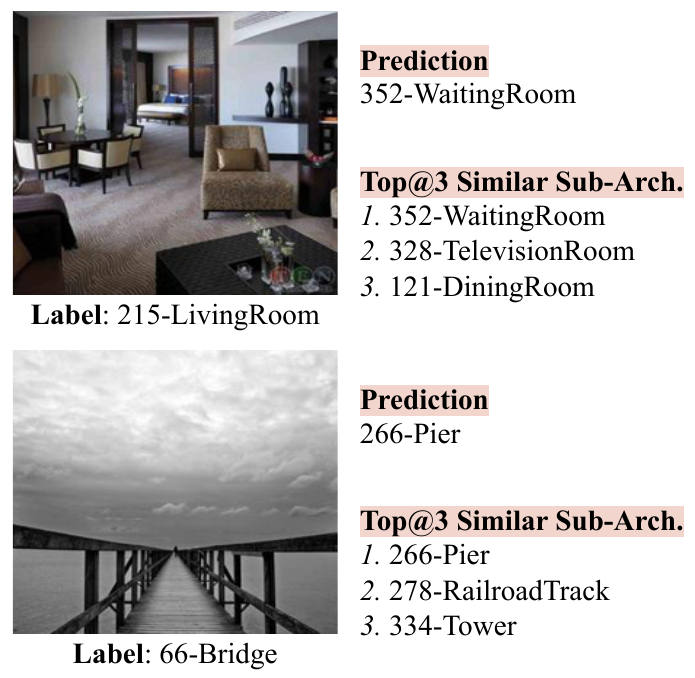}
 \label{fig5-6}
 \end{subfigure}
 \begin{subfigure}{0.24\textwidth}
 \centering
 \caption{DenseNet121 on Place365} \vspace{-2mm}
 \includegraphics[width=1.62in]{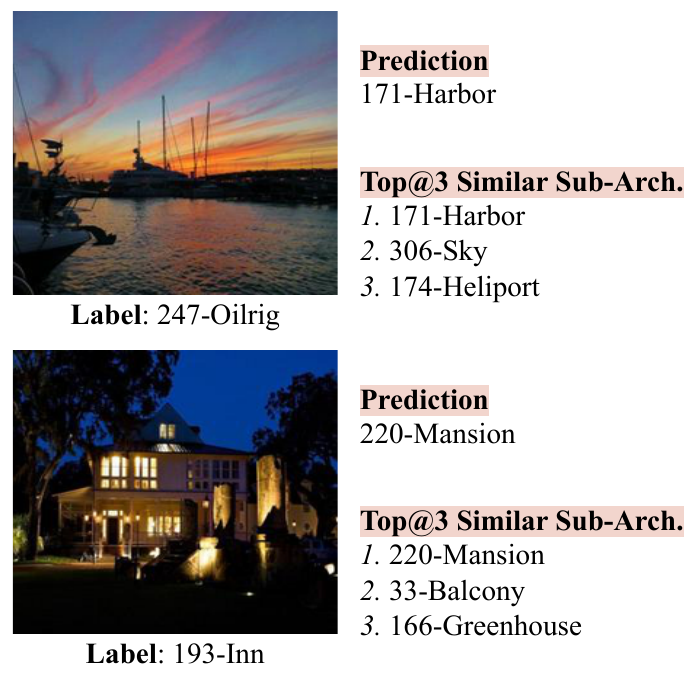}
 \label{fig5-7}
 \end{subfigure}
 \begin{subfigure}{0.24\textwidth}
 \centering
 \caption{DARTS-Net on Place365} \vspace{-2mm}
 \includegraphics[width=1.62in]{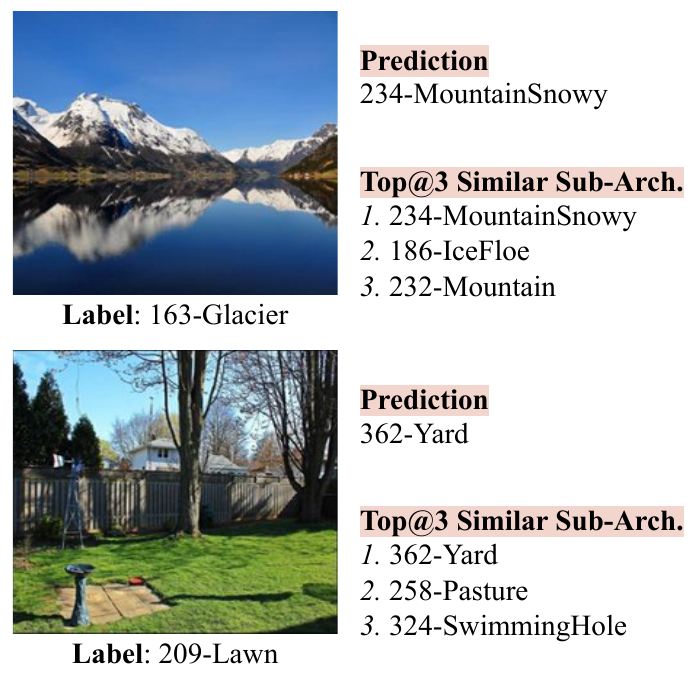}
 \label{fig5-8}
 \end{subfigure} \vspace{-4mm}
 \caption{
 Examples of misclassified classes with their Top@1 classification predictions and the `label-class' of the Top@3 similar sub-architectures with the correct labels.
 }%\vspace{-3mm}
 \label{fig5}
\end{figure*}

For VGG16 in Fig.~\ref{fig4-1}, the hit rate begins to increase from the 8th layer and reaches the top point in the last two layers.
We also find that the classes `Goldfish' and `AutoFactory' are gradually selected among the activated object parts in the visualization.
This supports our claim that the disentanglement in VGG16 starts in the middle layers, and the classes are selected in the top layers.
However, in the architectures with skip-connections, the results become far more complicated than VGG16, which has rarely been discussed in previous studies.
For ResNet50 in Fig.~\ref{fig4-2}, the highest hit rate is not in the last layer but the 16th layer.
The 12th, 13th, and 14th layers also get high hit rates.
From the visualization, we also find that the classes `Goldfish' and `AutoFactory' are selected before the last layer, and the best hit rate correspondingly emerges in the 16th layer.
For DenseNet121 in Fig.~\ref{fig4-3}, the hit rate gradually increases, but the values are not high in all layers.
This could result from the dense skip-connection severely mixing up the information from different layers.
The most complex results come from the automatically-searched architecture DARTS-Net in Fig.~\ref{fig4-4}.
It yields different results on the two datasets.
On ImageNet, the hit rate is high in the 12th and 15th layers, and the class `Goldfish' is selected in the corresponding layers.
However, the results are not good on Place365.
We believe this is because the architecture of DARTS-Net is searched on the CIFAR10 dataset, which has similar classes to ImageNet but a large domain gap with Place365.

In summary, we find that the skip-connections in ResNet50 and DARTS-Net can make the disentanglement end early, while the dense skip-connection in DenseNet121 severely mixes information up.
Specifically, the high-level semantic information can be extracted in the middle layers and sent to the top layers by the skip-connections in ResNet50 and DARTS-Net, which is quite different from what occurs in the VGG16 architecture with direct connections.
Meanwhile, the dense skip-connection in DenseNet121 severely amortizes the information in every layer, making it perform similarly to VGG16 but more challenging to disentangle.

\vspace{-0.2cm}
\subsubsection{How Does the Disentanglement Occur?}
%Combining the above results, w
We now study how the disentanglement occurs and investigate the inference process in DNNs.
Generally, different networks have different connection types, causing different inference processes.
Intuitively, direct connections successively transmit information layer by layer, while skip-connections amortize the information over all layers.
In VGG16 with direct connections, the patterns start to group into concrete semantics in the middle layers, \ie, the 8th layer to 10th layer, and the combined semantics are disentangled in the top layers, \ie, the 11th layer to 13th layer.
Such a process also emerges in the middle layers of ResNet50, \ie, the 10th layer to the 16th layer.
The difference is that the disentanglement skips over specific layers, such as the 15th, and ends in the 16th instead of the last layer.
Early disentanglement stopping can also be found in DARTS-Net, tested on the ImageNet dataset, where the semantic concepts have already been selected in the 12th and 15th layers.
By observing the visualization of the disentangled sub-architectures and the original architecture in DenseNet121, we find that some layers activate less valuable patterns in the input images for classification, \eg, the 3rd layer and the 4th layer. However, the high-level information is still extracted in the last layer.
This indicates that the information is severely mixed up in DenseNet121, where useful information for classification is amortized into each layer by dense skip-connections.

\subsection{Why DNNs Misjudge}
We try to explain why DNNs sometimes give wrong predictions.
To this end, we show the misclassified labels of the input images, and compute the average similarity between the sub-architectures of the misclassified labels and the correct ones.
For comparison, we also calculate the average similarity between the sub-architectures of the correct labels and all other labels except the misclassified ones.
%(\ie, depicting by the average similarity between the class's sub-architecture and the misclassified classes' sub-architectures) (\ie, depicting by the average similarity between this concept's sub-architecture and the other concepts' sub-architectures)
Given one class, the above experiment compares the sub-architecture similarity with the misclassified classes and other classes.
The results, shown in Table~\ref{tab1}, indicate that the misclassified input images tend to be assigned to classes with similar sub-architectures to the correct ones.
Fig.~\ref{fig5} shows some examples of misclassified input images with their classification predictions and similar concepts of the sub-architectures.
The sub-architectures tend to have similar semantic meanings to the target ones, \eg, for the label ``bulbul'', the Top@3 similar sub-architectures are ``Coucal'', ``Robin'' and ``Jay'', which are all birds.
\section{Conclusion}
\label{Conclusion}
In this paper, we introduce neural architecture disentanglement (NAD) for better understanding DNNs.
Starting from the current line of research, which aligns concepts to DNNs' units or layers, we try to link concepts to DNNs' sub-architectures instead.
Based on information theory, NAD learns to disentangle a pre-trained network in terms of tasks, forming information flows that describe the inference processes throughout the network.
%both containing different connection types,
We investigate the properties of NAD on object-based and scene-based datasets with DNNs ranging from handcrafted to automatically-searched.
% for DNNs 
The experimental results yield three new findings to explain the inner workings of DNNs, and an explanation of the classification performance is further discussed from the perspective of NAD.
%and provide fresh insights for future studies in
We hope NAD will shed light on the inference processes in DNNs for understanding how DNNs work.
{\small
\bibliographystyle{ieee_fullname}
\bibliography{egbib}
}
% \end{document}

% ---------------------------
\begin{appendices}
\onecolumn

\section{Formula Derivations}
\subsection{Derivation of Eq.~3}
According to the definition of mutual information, we expand the first term of Eq.~2 under the joint distribution of input $x^c$ and representation $r_{n-1}^c$:
%; 
\begin{equation}
\begin{split}
\label{eq17}
\mathcal{I}(x^c;r_{n-1}^c) &= \int\int P(r_{n-1}^c, x^c)\log \frac{P(r_{n-1}^c|x^c)}{P(r_{n-1}^c)}dx^cdr_{n-1}^c\\
&=\int\int P(r_{n-1}^c,x^c)\log P(r_{n-1}^c|x^c)dx^cdr_{n-1}^c - \int\int P(x^c|r_{n-1}^c)P(r_{n-1}^c)\log P(r_{n-1}^c)dx^cdr_{n-1}^c \\
&=\int\int P(r_{n-1}^c,x^c)\log P(r_{n-1}^c|x^c)dx^cdr_{n-1}^c - \int P(r_{n-1}^c)\log P(r_{n-1}^c)dr_{n-1}^c.
\end{split}
\end{equation}
Let $Q(r_{n-1}^c)$ be a variational approximation of $P(r_{n-1}^c)$, we have:
\begin{equation}
\begin{split}
\label{eq18}
KL\big[P(r_{n-1}^c)||Q(r_{n-1}^c)\big] \ge 0 \Rightarrow \int P(r_{n-1}^c)\log P(r_{n-1}^c)dr_{n-1}^c \ge \int P(r_{n-1}^c)\log Q(r_{n-1}^c)dr_{n-1}^c.
\end{split}
\end{equation}
Therefore, the trackable upper bound after applying the variational approximation is:
\begin{equation}
\begin{split}
\label{eq19}
\mathcal{I}(r_{n-1}^c; x^c) \le& \int\int P(r_{n-1}^c|x^c)P(x^c)\log \frac{P(r_{n-1}^c|x^c)}{Q(x^c)}dx^cdr_{n-1}^c \\
=&\mathbb{E}_{x^c\sim P(x^c)}\Big[KL\big[P(r_{n-1}^c|x^c)||Q(r_{n-1}^c)\big]\Big].
\end{split}
\end{equation}

For the second term of Eq.~2, we expand it as the joint distribution of representation $r_{n-1}^c$ and the target $y^c$:
\begin{equation}
\begin{split}
\label{eq20}
\mathcal{I}(r_{n-1}^c; y^c) &= \int\int P(r_{n-1}^c,y^c)\log \frac{P(y^c|r_{n-1}^c)}{P(y^c)}dr_{n-1}^cdy^c \\
&= \int\int P(r_{n-1}^c,y^c)\log P(y^c|r_{n-1}^c)dy^cdr_{n-1}^c - \int P(y^c)\log P(y^c)dy^c \\
&= \int\int P(r_{n-1}^c,y^c)\log P(y^c|r_{n-1}^c)dy^cdr_{n-1}^c + \mathcal{H}(y^c) \\
&\ge \int\int P(y^c|r_{n-1}^c)P(r_{n-1}^c) \log P(y^c|r_{n-1}^c)dy^cdr_{n-1}^c,
\end{split}
\end{equation}
where $\mathcal{H}(y^c) \ge 0$ is the information entropy of $y^c$.
Let $Q(y^c|r_{n-1}^c)$ be a variational approximation of $P(y^c|r_{n-1}^c)$, we have:
\begin{equation}
\begin{split}
\label{eq21}
KL\big[P(y^c|r_{n-1}^c) || Q(y^c|r_{n-1}^c)\big] \ge 0 \Rightarrow \int P(y^c|r_{n-1}^c)\log P(y^c|r_{n-1}^c)dy^c \ge \int P(y^c|r_{n-1}^c)\log Q(y^c|r_{n-1}^c)dy^c.
\end{split}
\end{equation}
By applying the variational approximation, the trackable lower bound of the mutual information between $r_{n-1}^c$ and $y^c$ is:
\begin{equation}
\begin{split}
\label{eq22}
\mathcal{I}(r_{n-1}^c;y^c)\ge\int\int P(r_{n-1}^c,y^c) \log Q(y^c|r_{n-1}^c) dy^cdr_{n-1}^c.
\end{split}
\end{equation}
Assuming that the representation $r_{n-1}^c$ is independent of the label $y^c$, \emph{i.e.}, $P(r_{n-1}^c|x^c, y^c) = P(r_{n-1}^c|x^c)$, we have:
\begin{equation}
\begin{split}
\label{eq23}
P(x^c,r_{n-1}^c,y^c)=P(r_{n-1}^c|x^c,y^c)P(y^c|x^c)P(x^c) =P(r_{n-1}^c|x^c)P(y^c|x^c)P(x^c).
\end{split}
\end{equation}
Then, the joint distribution of $r_{n-1}^c$ and $y^c$ can be written as:
\begin{equation}
\begin{split}
\label{eq24}
P(r_{n-1}^c,y^c)=\int P(x^c,r_{n-1}^c,y^c)dx^c =\int P(r_{n-1}^c|x^c)P(y^c|x^c)P(x^c)dx^c.
\end{split}
\end{equation}
Combining Eq.~22 with Eq.~24, we get the lower bound:
\begin{equation}
\begin{split}
\label{eq25}
\mathcal{I}(r_{n-1}^c;y^c)&\ge\int\int\int P(x^c)P(r_{n-1}^c|x^c)P(y^c|x^c)\log Q(y^c|r_{n-1}^c)dy^cdr_{n-1}^cdx^c \\
&=\mathbb{E}_{x^c\sim P(x^c)}\Big[\mathbb{E}_{r_{n-1}^c\sim P(r_{n-1}^c|x^c)}\big[\int P(y^c|x^c)\log Q(y^c|r_{n-1}^c)dy^c\big]\Big] \\
&=\mathbb{E}_{x^c\sim P(x^c)}\Big[\mathbb{E}_{r_{n-1}^c\sim P(r_{n-1}^c|x^c)}\big[\log Q(y^c|r_{n-1}^c)\big]\Big].
\end{split}
\end{equation}
With the Eq.~19 and Eq.~25, the variational upper bound of Eq.~2 is derived to Eq.~3 as:
\begin{equation}
\begin{split}
\label{eq26}
\mathcal{\widetilde{L}}_{IB}=\mathbb{E}_{x^c\sim P(x^c)}\Big[\beta KL\big[P(r_{n-1}^c|x^c)||Q(r_{n-1}^c)\big]
-\mathbb{E}_{r_{n-1}^c\sim P(r_{n-1}^c|x^c)}\big[\log {Q(y^c|r_{n-1}^c)
}\big]\Big].
\end{split}
\end{equation}
The derivation of Eq.~11 is the same as the derivation of the second term of Eq.~3. 
\subsection{Derivation of Eq.~9 and Eq.~13}
For the first term of Eq.~6, the KL divergence can be expanded with Eq.~8.
Taking the univariate Gaussian distribution as example, the KL divergence is:
\begin{equation}
\begin{split}
\label{eq27} % KL\big[P(r_{n-1}^c|x^c)||Q(r_{n-1}^c)\big]=
KL\Big[ \mathcal{N}(0,1)||\mathcal{N}\big(r_i^c\cdot \mu^c_i, (r_i^c\cdot \sigma^c_i)^2\big) \Big] =& \int \frac{1}{\sqrt{2\pi}r_i^c\sigma^c_i}e^{-(x-r_i^c\mu^c_i)^2/2(r_i^c\sigma^c_i)^2}\big(\log \frac{e^{-(x-r_i^c\mu^c_i)^2/2(r_i^c\sigma^c_i)^2}}{r_i^c\sigma^c_ie^{-x^2/2}}\big)dx \\
=& \frac{1}{2}\int \frac{1}{\sqrt{2\pi}r_i^c\sigma^c_i}e^{-(x-r_i^c\mu^c_i)^2/2(r_i^c\sigma^c_i)^2}\big(x^2-\log (r_i^c\sigma^c_i)^2-(x-r_i^c\mu^c_i)^2/(r_i^c\sigma^c_i)^2 \big)dx \\
=& \frac{1}{2}\big((r_i^c\cdot \sigma^c_i)^2-\log{(r_i^c\cdot \sigma^c_i)^2}+(r_i^c\cdot \mu^c_i)^2-1\big).
\end{split}
\end{equation}
For the second term of Eq.~6, the log-likelihood function of the univariate Gaussian distribution is:
\begin{equation}
\begin{split}
\label{eq28}
-\log Q(\widetilde{r}_i^c|r_i^c) =& -\log \frac{1}{\sqrt{2\pi}r_i^c\sigma^c_i}e^{-(r_i^c-r_i^c\mu^c_i)^2/2(r_i^c\sigma^c_i)^2} \\
=& \frac{1}{2}\big(||r^c_i - \mu^c_i\cdot r^c_i||_2^2 + \log 2\pi + \log{(r_i^c\cdot \sigma^c_i)^2}\big).
\end{split}
\end{equation}
Combining Eq.~27 and Eq.~28, the constraint in Eq.~6 for the $i$-th hidden layer can be derived to Eq.~9 as:
\begin{equation}
\begin{split}
\label{eq29}
\mathcal{\widetilde{L}}^c_i&=\frac{\beta}{2}\big((r_i^c\cdot \sigma^c_i)^2-\log{(r_i^c\cdot \sigma^c_i)^2}+(r_i^c\cdot \mu^c_i)^2-1\big)\\
&+\frac{1}{2}\big(||r^c_i - \mu^c_i\cdot r^c_i||_2^2 + \log 2\pi + \log{(r_i^c\cdot \sigma^c_i)^2}\big).
\end{split}
\end{equation}
After reducing the noisy level by fixing $\epsilon_i$ in Eq.~7 to its mean value, \ie, $0$, $\sigma^c_i$ is freed to be any value that does not affect the optimization process.
Therefore, Eq.~29 can be simplified to Eq.~13 as:
\begin{equation}
\begin{split}
\label{eq30}
\mathcal{\widetilde{L}}^c_i=\beta(r_i^c\cdot \mu^c_i)^2
+||r^c_i - \mu^c_i\cdot r^c_i||_2^2.
\end{split}
\end{equation}
\subsection{Derivation of Eq.~12}
For Eq.~11, the log-likelihood function of the univariate Bernoulli distribution can be directly written as:
\begin{equation}
\begin{split}
\label{eq31}
-\log Q(y^c|\widetilde{y}^c) &= -\log \big(f_n(r^c_{n-1})\big)^{y^c}\big(1-f_n(r^c_{n-1})\big)^{1-y^c} \\
&= -y^c\log f_n(r^c_{n-1}) -(1-y^c)\log\big(1-f_n(r^c_{n-1})\big)\\
&=-y^c\log\widetilde{y}^c-(1-y^c)\log(1-\widetilde{y}^c).
\end{split}
\end{equation}
\section{Hyper-Parameter $\beta$}
\begin{figure}[t]
\centering
\begin{subfigure}{0.495\textwidth}
\centering
\caption{Results of VGG16 on ImageNet} \vspace{-0.2cm}
\includegraphics[width=0.99\linewidth]{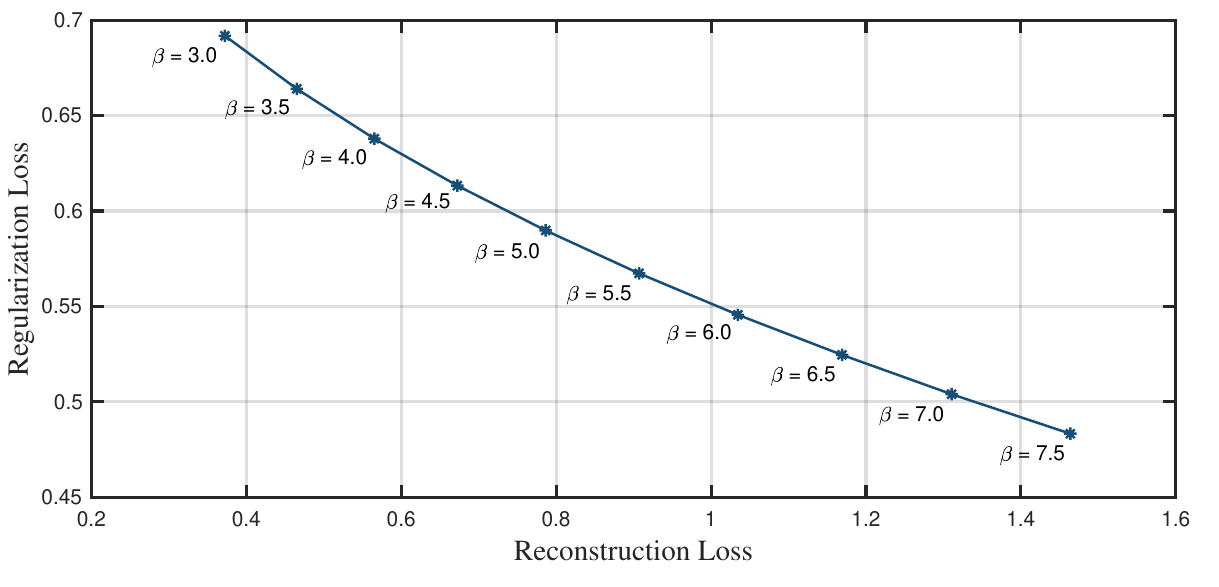}
\label{fig6-1}
\end{subfigure}
\begin{subfigure}{0.495\textwidth}
\centering
\caption{Results of ResNet50 on ImageNet} \vspace{-0.2cm}
\includegraphics[width=0.99\linewidth]{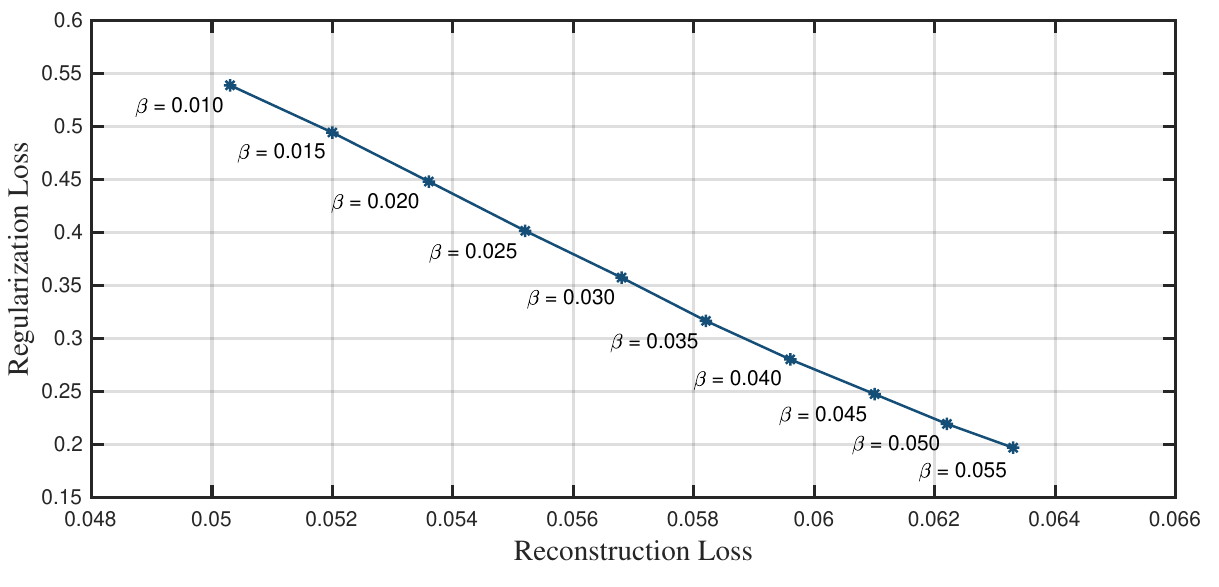}
\label{fig6-2}
\end{subfigure}
\begin{subfigure}{0.495\textwidth}
\centering
\caption{Results of DenseNet121 on ImageNet} \vspace{-0.2cm}
\includegraphics[width=0.99\linewidth]{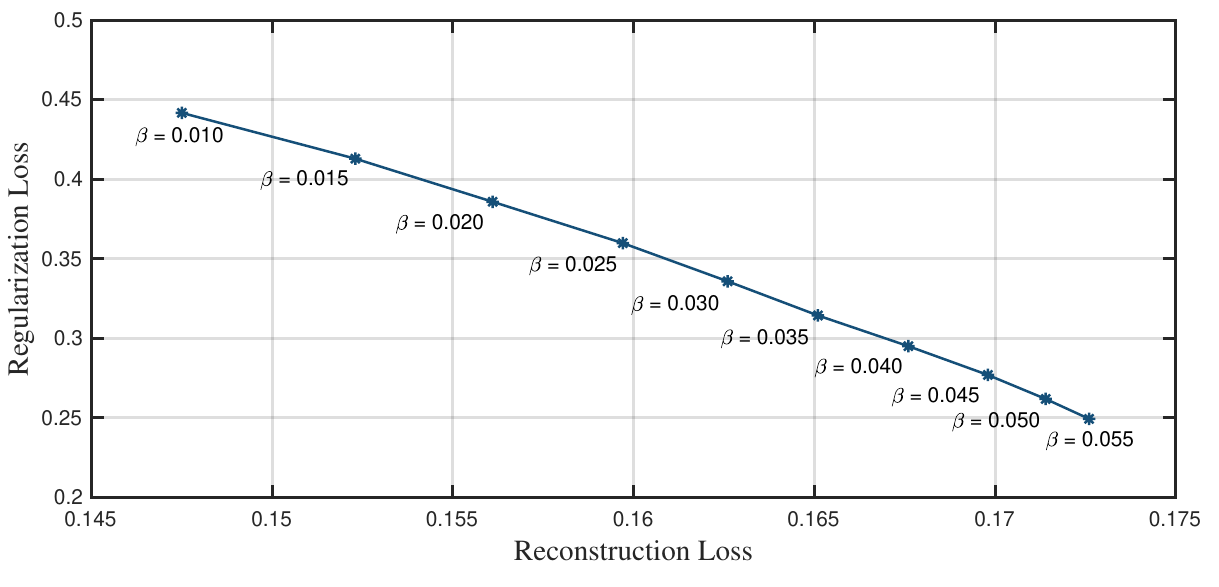} 
\label{fig6-3}
\end{subfigure}
\begin{subfigure}{0.495\textwidth}
\centering
\caption{Results of DARTS-Net on ImageNet} \vspace{-0.2cm}
\includegraphics[width=0.99\linewidth]{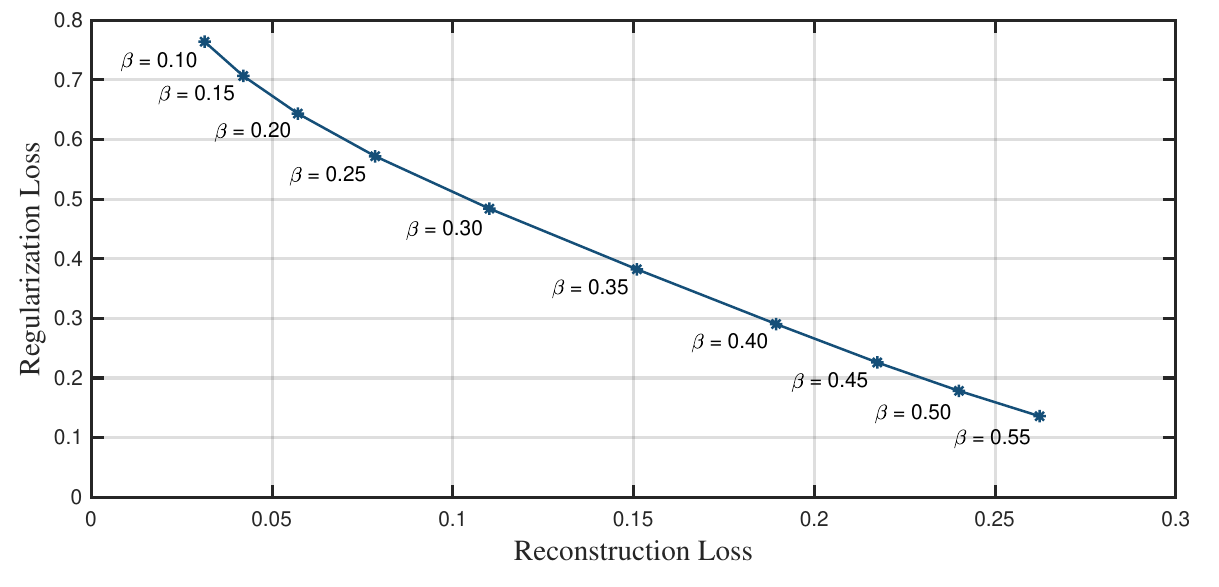} 
\label{fig6-4}
\end{subfigure}
\begin{subfigure}{0.495\textwidth}
\centering
\caption{Results of VGG16 on Place365} \vspace{-0.2cm}
\includegraphics[width=0.99\linewidth]{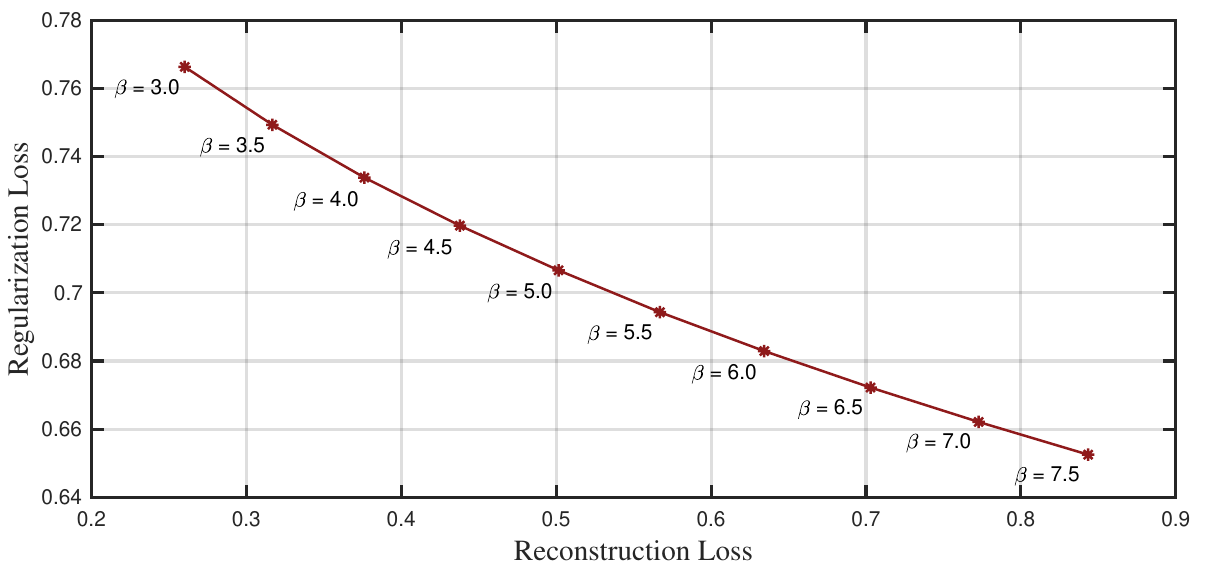} 
\label{fig6-5}
\end{subfigure}
\begin{subfigure}{0.495\textwidth}
\centering
\caption{Results of ResNet50 on Place365} \vspace{-0.2cm}
\includegraphics[width=0.99\linewidth]{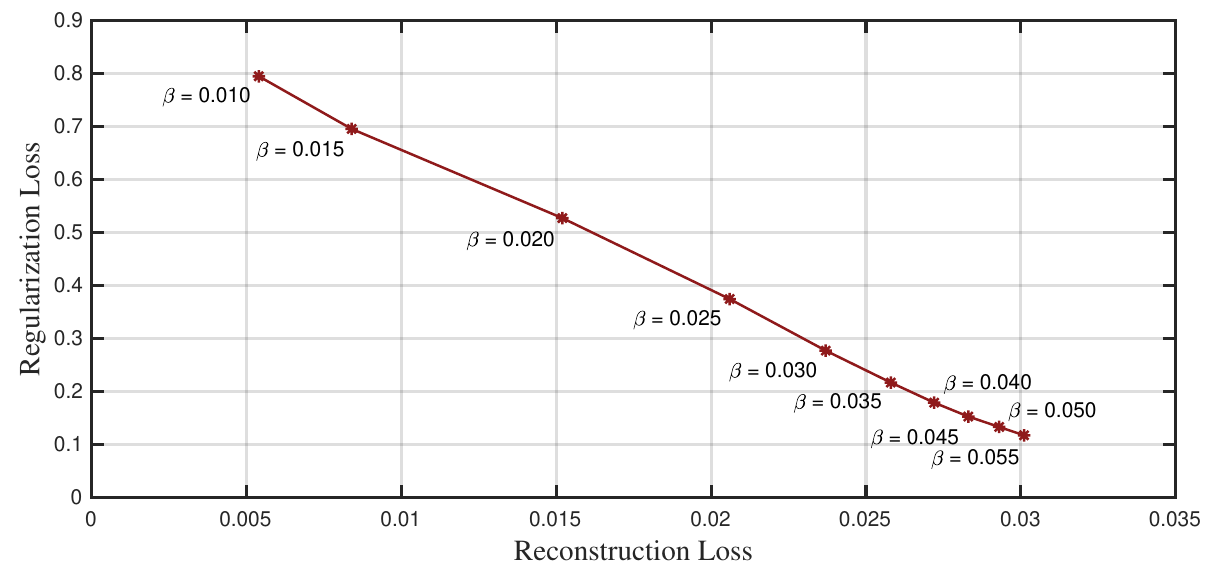} 
\label{fig6-6}
\end{subfigure}
\begin{subfigure}{0.495\textwidth}
\centering
\caption{Results of DenseNet121 on Place365} \vspace{-0.2cm}
\includegraphics[width=0.99\linewidth]{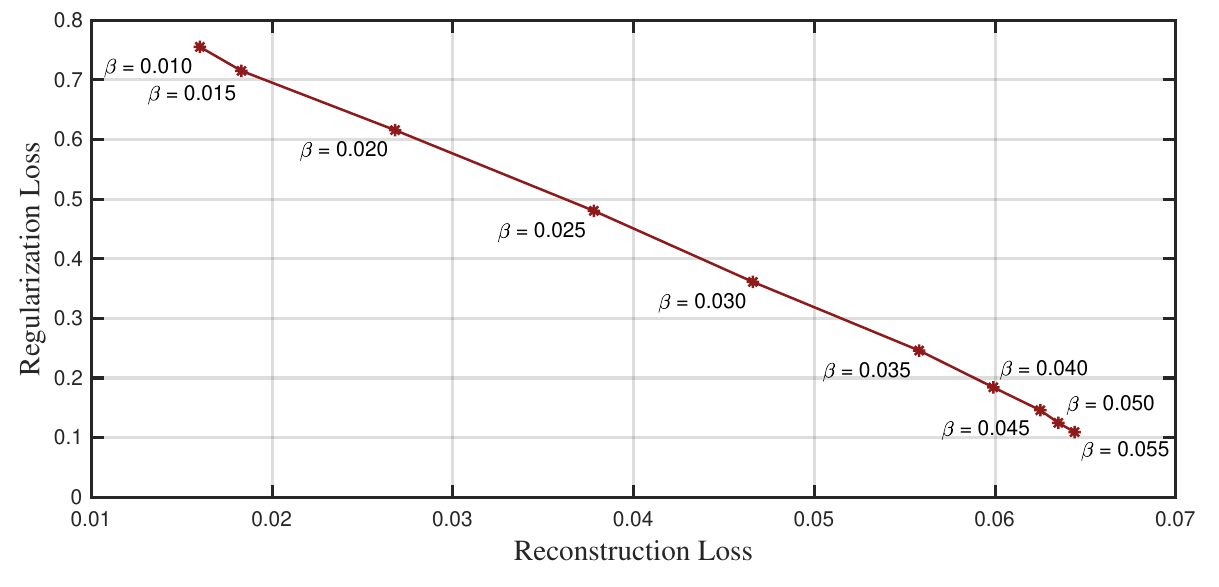} 
\label{fig6-7}
\end{subfigure}
\begin{subfigure}{0.495\textwidth}
\centering
\caption{Results of DARTS-Net on Place365} \vspace{-0.2cm}
\includegraphics[width=0.99\linewidth]{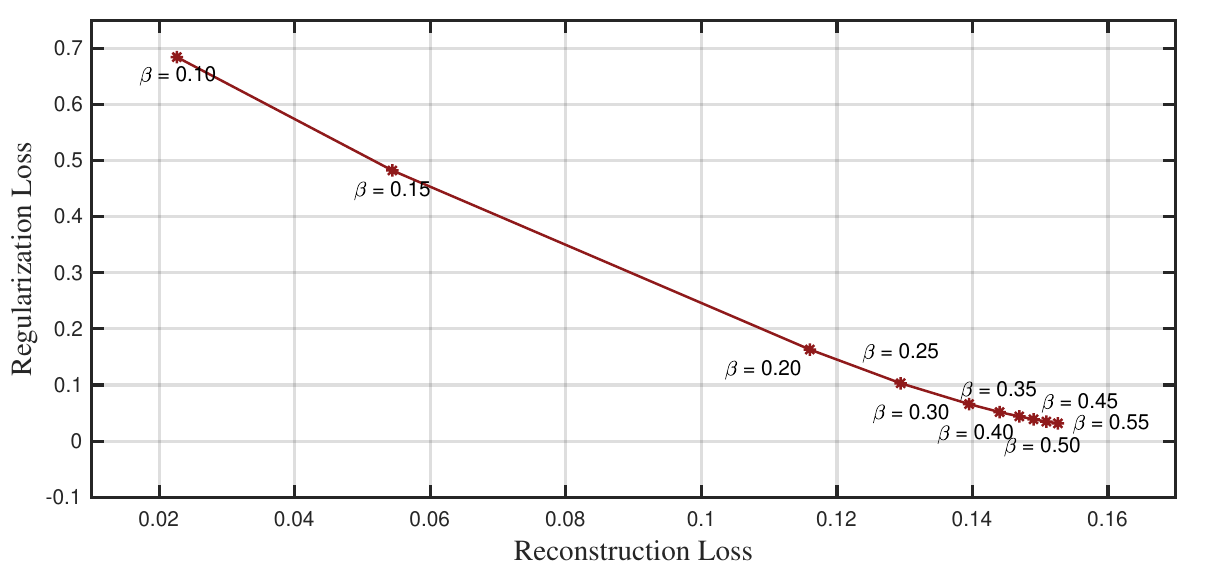} 
\label{fig6-8}
\end{subfigure}
\caption{Reconstruction loss vs. regularization loss with different hyper-parameter $\beta$.}
\label{fig6}
\end{figure}

We use the first $5\%$ labels for determining the hyper-parameter $\beta$ according the Elbow method, with which the reconstruction loss and regularization loss are balanced.
For VGG16 in Figs.~\ref{fig6-1} and~\ref{fig6-5}, the hyper-parameter $\beta=4.5$ and $\beta=5.0$ balance the losses well on ImageNet and Place365, respectively.
For ResNet50 in Figs.~\ref{fig6-2} and~\ref{fig6-6}, the hyper-parameter $\beta=0.02$ and $\beta=0.02$ balance the losses well on ImageNet and Place365, respectively.
For DenseNet121 in Figs.~\ref{fig6-3} and~\ref{fig6-7}, the hyper-parameter $\beta=0.02$ and $\beta=0.02$ balance the losses well on ImageNet and Place365, respectively.
For DARTS-Net in Figs.~\ref{fig6-4} and~\ref{fig6-8}, the hyper-parameter $\beta=0.45$ and $\beta=0.20$ balance the losses well on ImageNet and Place365, respectively.
\section{More Visualization Examples}
More results of the visualization with the activated feature maps are shown from Fig.~7 to Fig.~16.
\begin{figure}[t]
\centering
\includegraphics[width=1\linewidth]{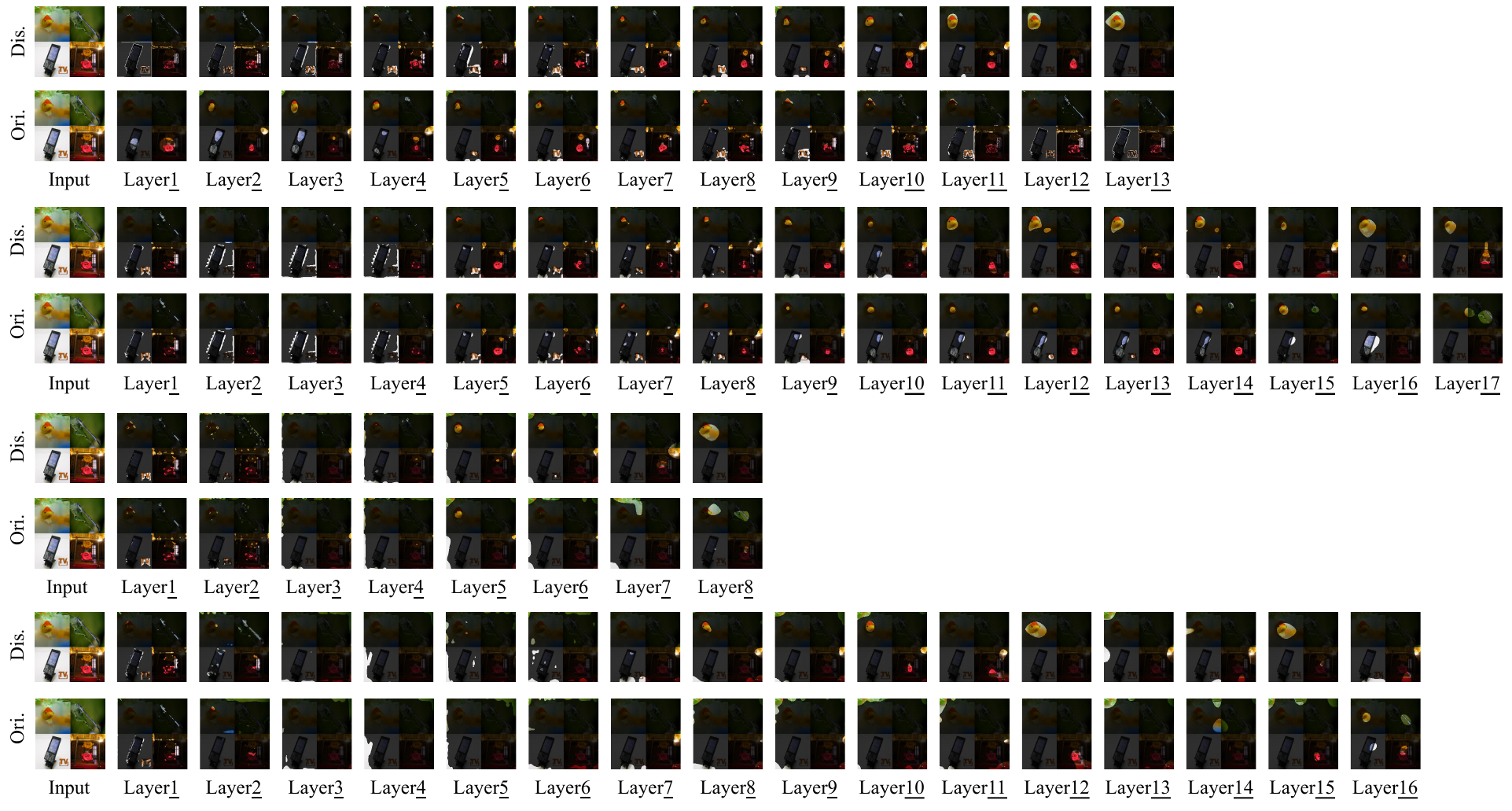}
\caption{
Example combining the images with `label-concept': `1-Goldfish', `320-Damselfly', `487-MobilePhone', and `489-ChainLinkFence' from the validation set of ImageNet.
The results from top to bottom are from VGG16, ResNet50, DenseNet121, and DARTS-Net, respectively.
}
\label{fig7}
\end{figure}
\begin{figure}[t]
\centering
\includegraphics[width=1\linewidth]{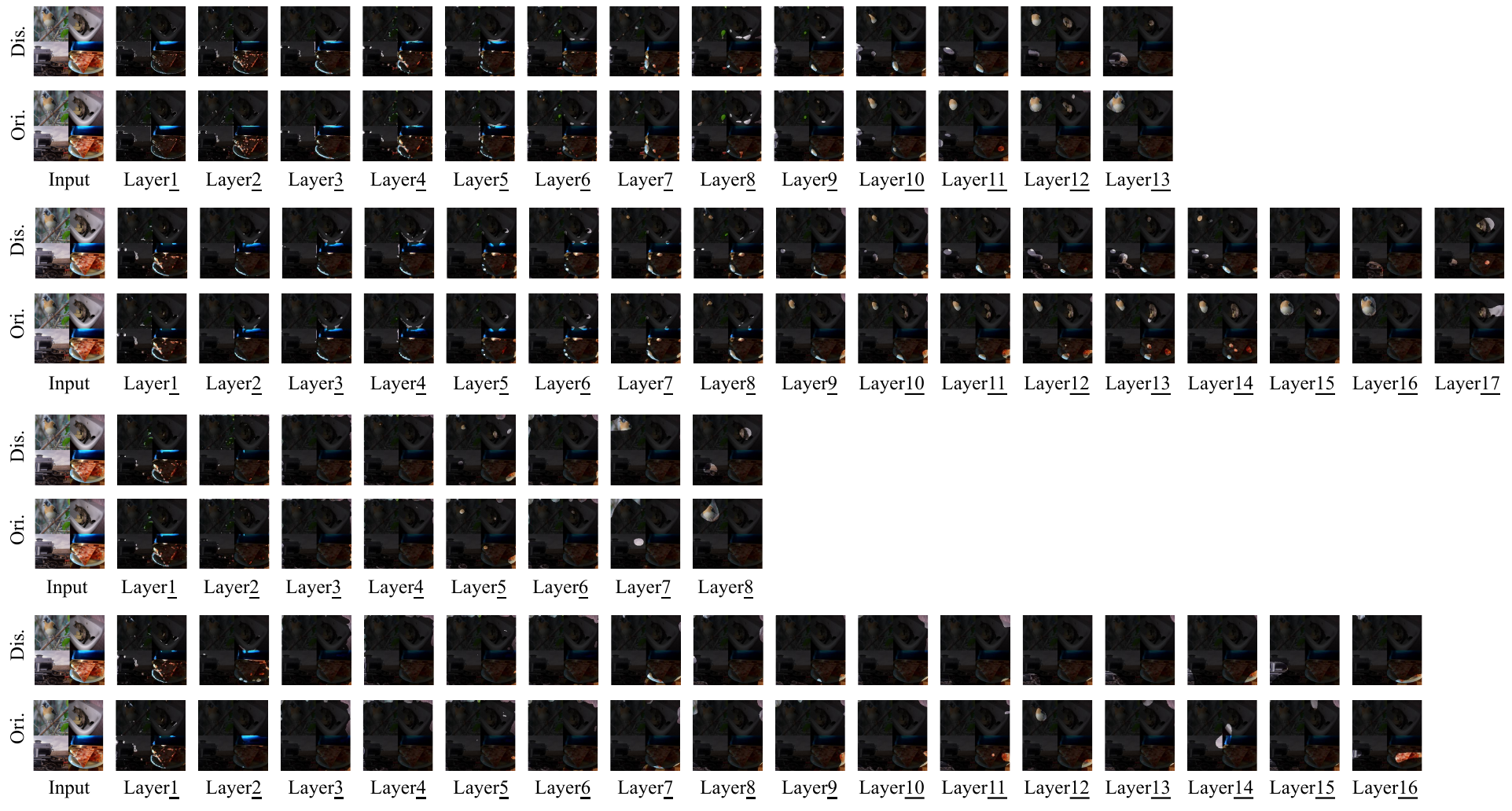}
\caption{
Example combining the images with `label-concept': `10-Goldfinch', `896-Washer', `820-SteelArchBridge', and `963-Potpie' from the validation set of ImageNet.
The results from top to bottom are from VGG16, ResNet50, DenseNet121, and DARTS-Net, respectively.
}
\label{fig8}
\end{figure}
\begin{figure}[t]
\centering
\includegraphics[width=1\linewidth]{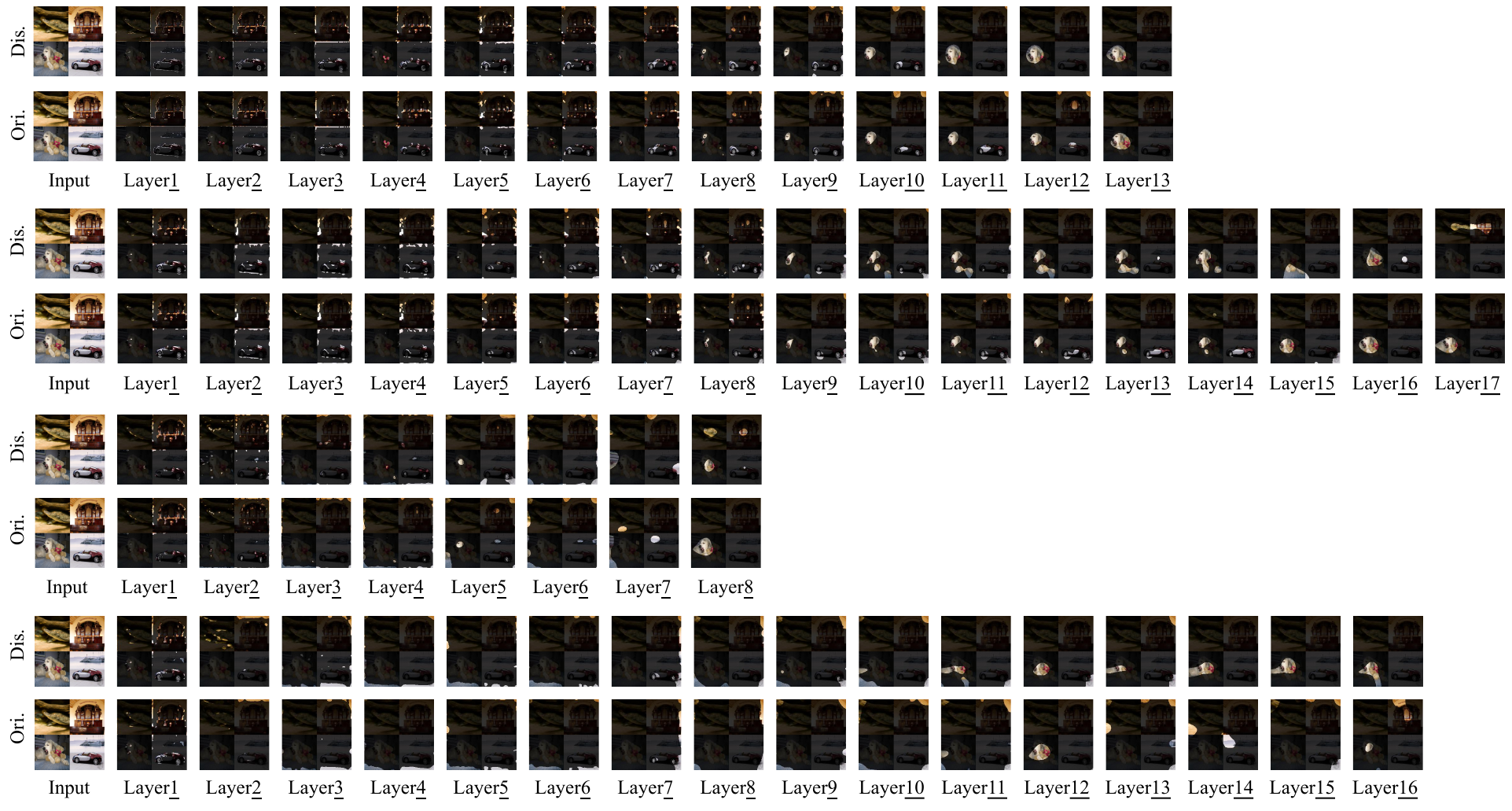}
\caption{
Example combining the images with `label-concept': `35-Terrapin', `687-Oscilloscope', `257-Samoyed', and `511-Corkscrew' from the validation set of ImageNet.
The results from top to bottom are from VGG16, ResNet50, DenseNet121, and DARTS-Net, respectively.
}
\label{fig9}
\end{figure}
\begin{figure}[t]
\centering
\includegraphics[width=1\linewidth]{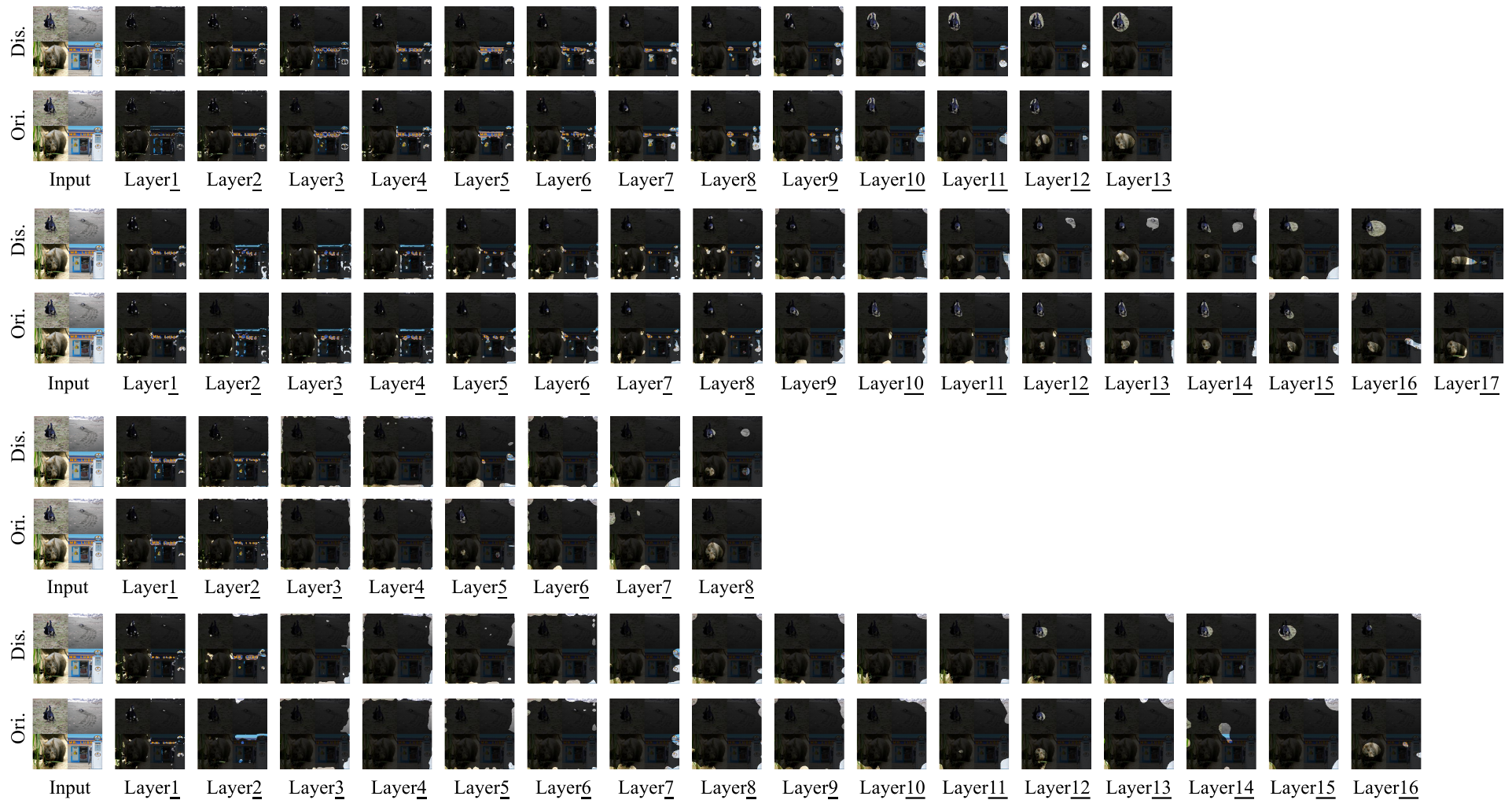}
\caption{
Example combining the images with `label-concept': `80-Ptarmigan', `145-Albatross', `106-jellyfish', and `571-Goblet' from the validation set of ImageNet.
The results from top to bottom are from VGG16, ResNet50, DenseNet121, and DARTS-Net, respectively.
}
\label{fig10}
\end{figure}
\begin{figure}[t]
\centering
\includegraphics[width=1\linewidth]{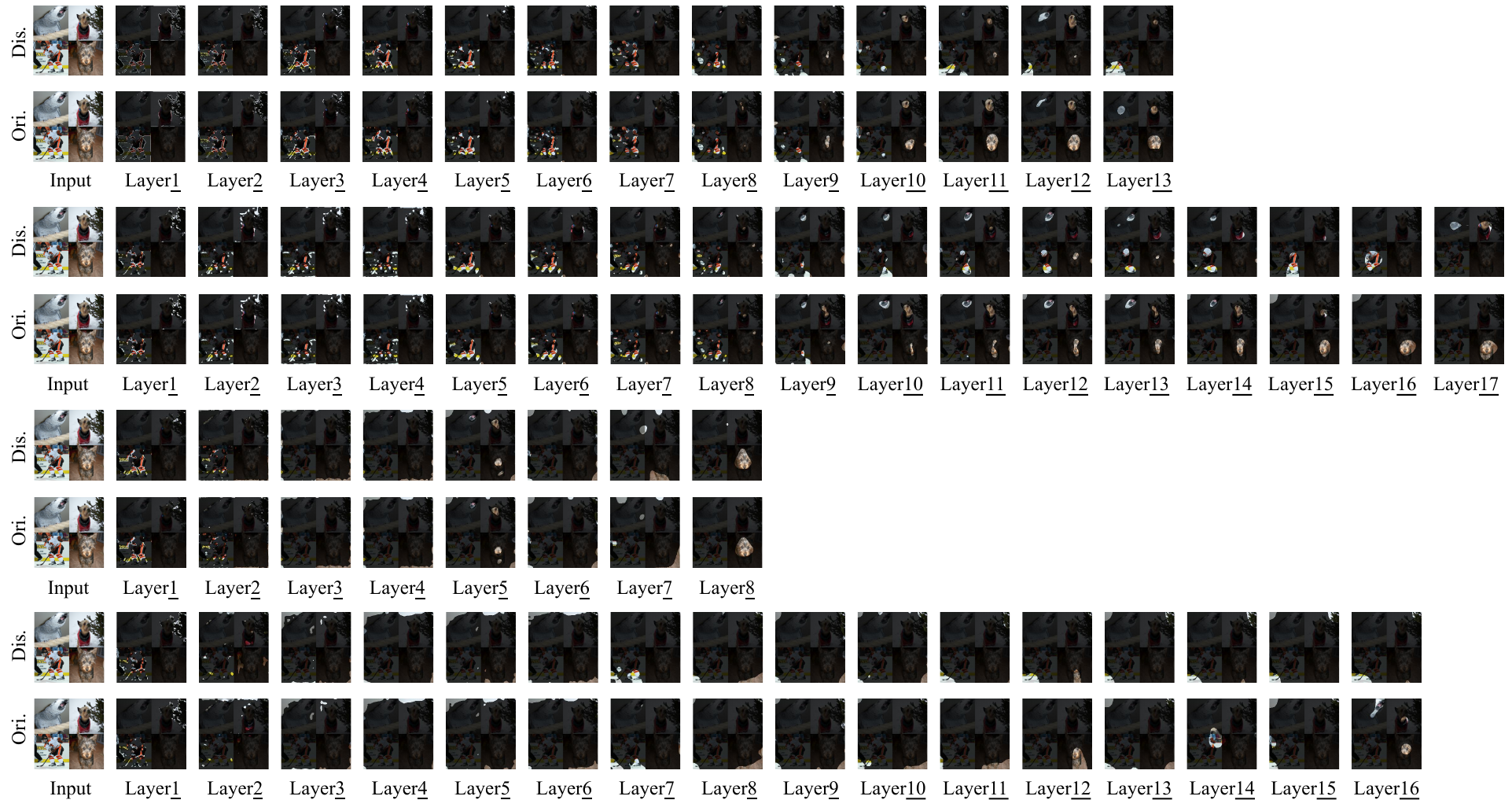}
\caption{
Example combining the images with `label-concept': `87-Macaw', `193-DandieDinmont', `746-PunchingBag', and `187-WireHairedFoxTerrier' from the validation set of ImageNet.
The results from top to bottom are from VGG16, ResNet50, DenseNet121, and DARTS-Net, respectively.
}
\label{fig11}
\end{figure}
\begin{figure}[t]
\centering
\includegraphics[width=1\linewidth]{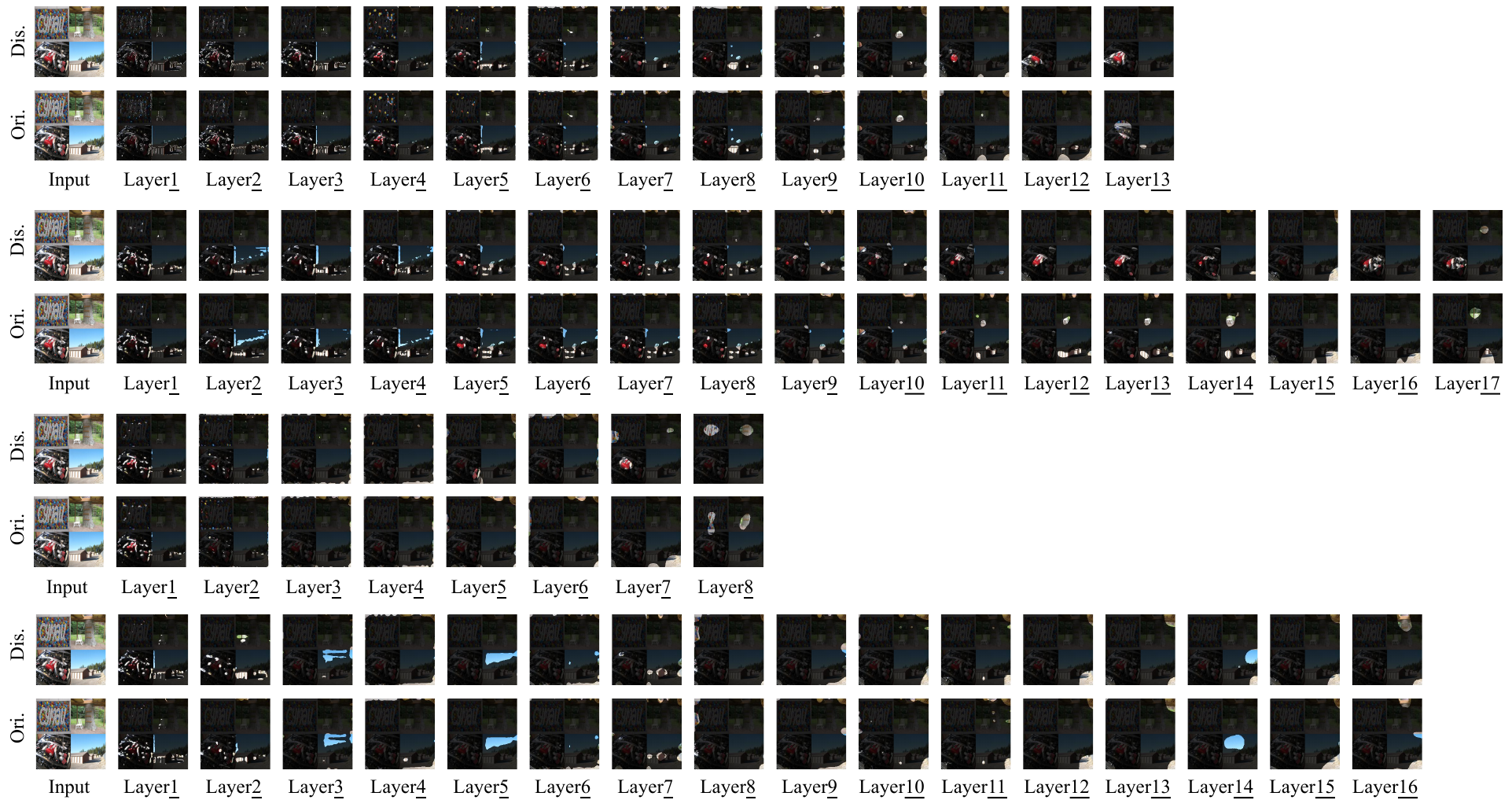}
\caption{
Example combining the images with `label-concept': `34-BallPit', `339-TreeHouse', `28-AutoFactory', and `221-ManufacturedHome' from the validation set of Place365.
The results from top to bottom are from VGG16, ResNet50, DenseNet121, and DARTS-Net, respectively.
}
\label{fig12}
\end{figure}
\begin{figure}[t]
\centering
\includegraphics[width=1\linewidth]{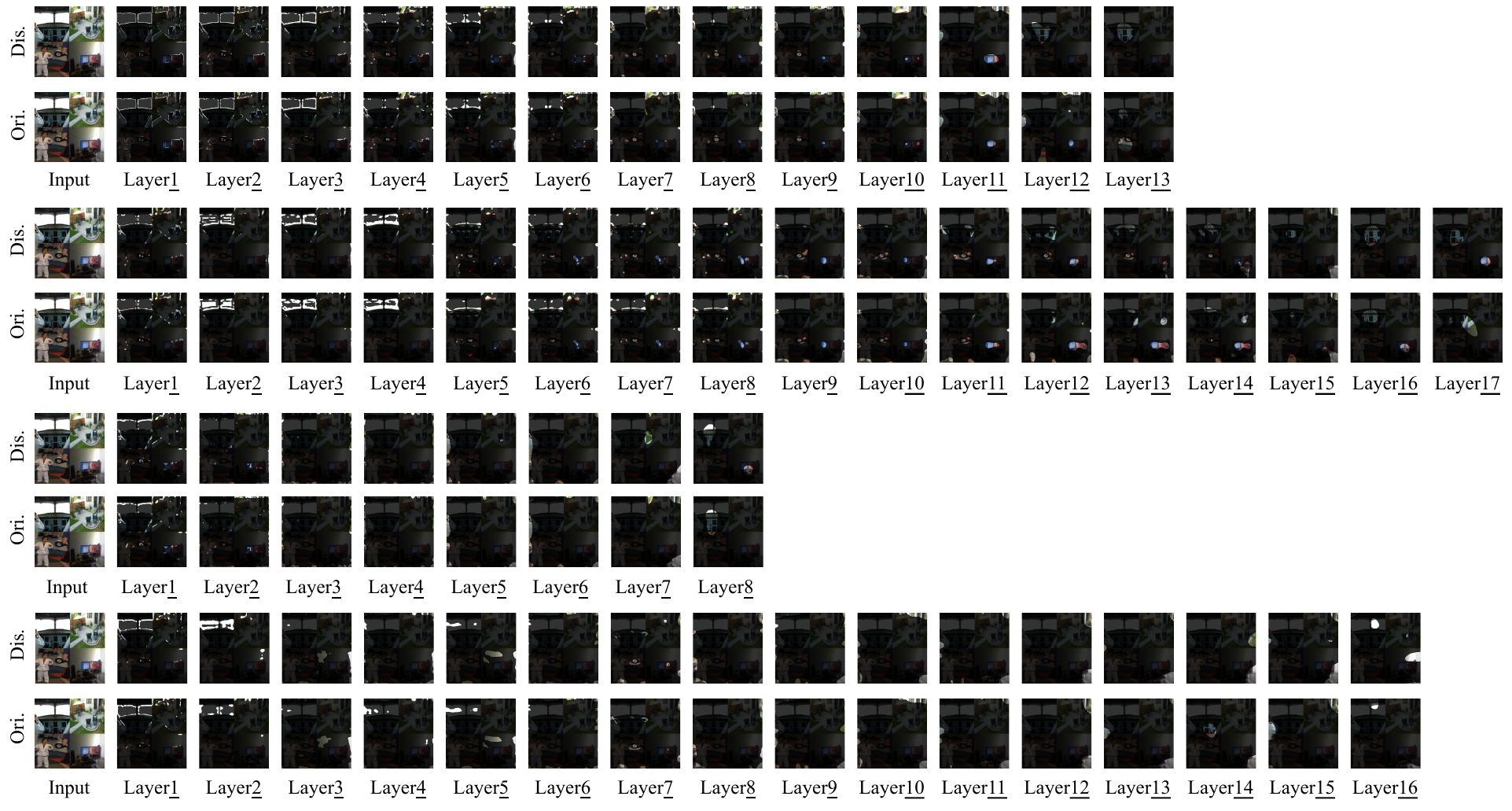}
\caption{
Example combining the images with `label-concept': `98-Cockpit', `259-Patio', `225-MartialArtsGym', and `100-ComputerRoom' from the validation set of Place365.
The results from top to bottom are from VGG16, ResNet50, DenseNet121, and DARTS-Net, respectively.
}
\label{fig13}
\end{figure}
\begin{figure}[t]
\centering
\includegraphics[width=1\linewidth]{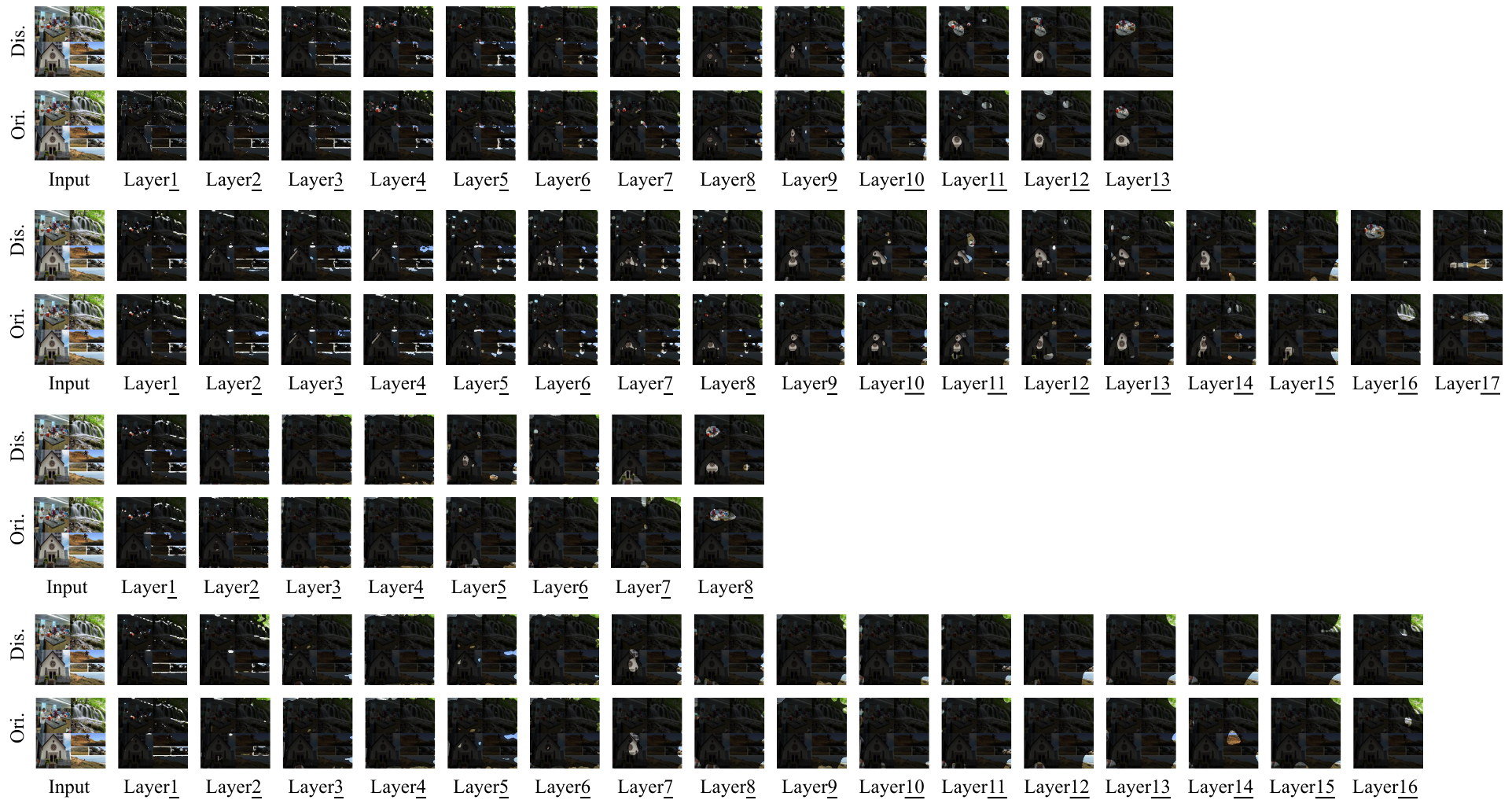}
\caption{
Example combining the images with `label-concept': `56-BiologyLaboratory', `355-Waterfall', `327-Synagogue', and `233-MountainPath' from the validation set of Place365.
The results from top to bottom are from VGG16, ResNet50, DenseNet121, and DARTS-Net, respectively.
}
\label{fig14}
\end{figure}
\begin{figure}[t]
\centering
\includegraphics[width=1\linewidth]{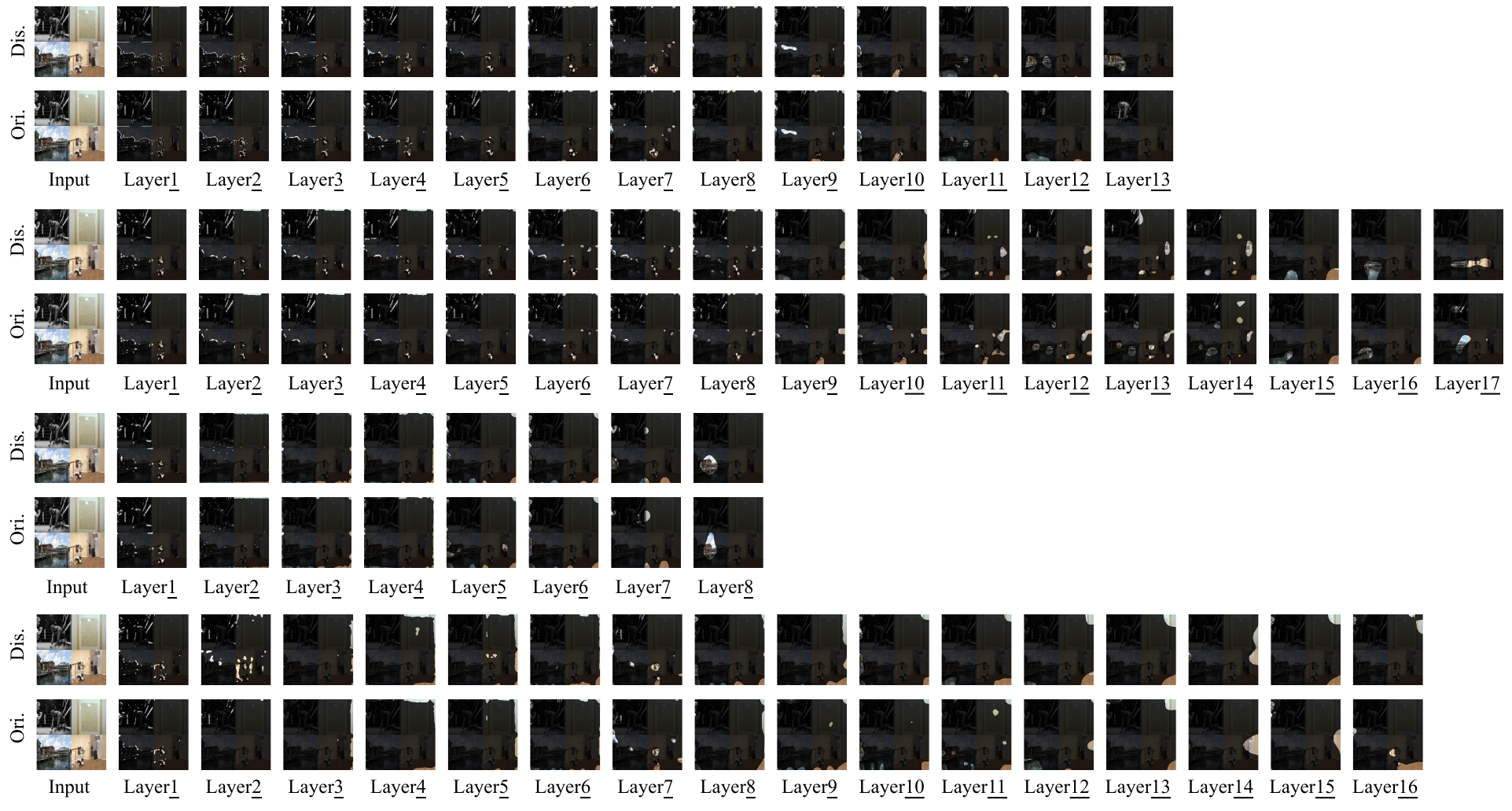}
\caption{
Example combining the images with `label-concept': `65-BoxingRing', `19-ArtGallery', `57-Boardwalk', and `168-Gymnasium' from the validation set of Place365.
The results from top to bottom are from VGG16, ResNet50, DenseNet121, and DARTS-Net, respectively.
}
\label{fig15}
\end{figure}
\begin{figure}[t]
\centering
\includegraphics[width=1\linewidth]{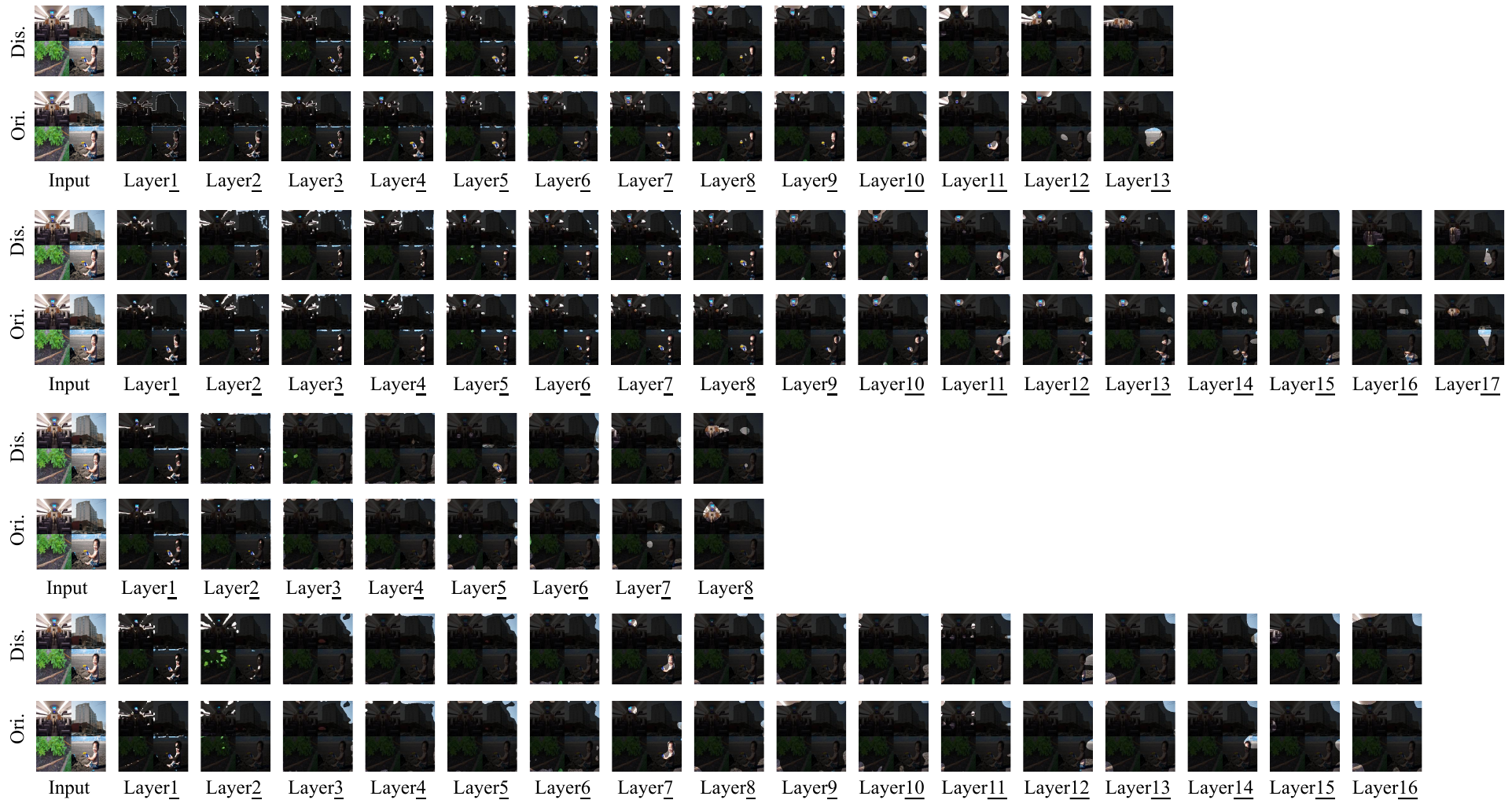}
\caption{
Example combining the images with `label-concept': `70-BusInterior', `8-ApartmentBuilding', `345-VegetableGarden', and `206-Landfill' from the validation set of Place365.
The results from top to bottom are from VGG16, ResNet50, DenseNet121, and DARTS-Net, respectively.
}
\label{fig16}
\end{figure}

\end{appendices}

\end{document}